\documentclass[journal]{IEEEtran}

\ifCLASSINFOpdf

\else

\fi

\usepackage[small]{caption}
\usepackage{graphicx}

\usepackage{amssymb}
\usepackage{pifont}
\newcommand{\cmark}{\ding{51}}%
\newcommand{\xmark}{\ding{55}}%

\begin{document}

\title{Long Short-Term Memory with Gate and State Level Fusion for Light Field-Based Face Recognition}

\author{Alireza~Sepas-Moghaddam,~\IEEEmembership{Member,~IEEE,}
        Ali~Etemad,~\IEEEmembership{Senior Member,~IEEE,}
        Fernando~Pereira,~\IEEEmembership{Fellow,~IEEE,}
        and~Paulo~Lobato Correia,~\IEEEmembership{Senior Member,~IEEE}


}

%
%

\markboth{Submitted to IEEE Transactions on Information Forensics and Security}%
{Shell \MakeLowercase{\textit{et al.}}: Bare Demo of IEEEtran.cls for IEEE Journals}
%



\maketitle

\begin{abstract}
Long Short-Term Memory (LSTM) is a prominent recurrent neural network for extracting dependencies from sequential data such as time-series and multi-view data, having achieved impressive results for different visual recognition tasks. A conventional LSTM network can learn a model to posteriorly extract information from one input sequence. However, if two or more dependent sequences of data are simultaneously acquired, the conventional LSTM networks may only process those sequences consecutively, not taking benefit of the information carried out by their mutual dependencies. 
In this context, this paper proposes two novel LSTM cell architectures that are able to jointly learn from multiple sequences simultaneously acquired, targeting to create richer and more effective models for recognition tasks. The efficacy of the novel LSTM cell architectures is assessed by integrating them into deep learning-based methods for face recognition with multi-view, light field images. The new cell architectures jointly learn the scene horizontal and vertical parallaxes available in a light field image, to capture richer spatio-angular information from both directions. A comprehensive evaluation, with the IST-EURECOM LFFD dataset using three challenging evaluation protocols, shows the advantage of using the novel LSTM cell architectures for face recognition over the state-of-the-art light field-based methods.
These results highlight the added value of the novel cell architectures when learning from correlated input sequences.

\end{abstract}


%
\IEEEpeerreviewmaketitle

\section{Introduction}

Recent years have witnessed rapid advances in the field of machine learning with the development of deep neural networks \cite{LeCun2,deepsurvey} and the emergence of powerful hardware resources, like graphics processing units (GPU) \cite{LeCun}. Nowadays, due to their superior representation and prediction performance, deep Convolutional Neural Networks (CNNs) are increasingly adopted for visual recognition and description tasks \cite{CNNSurvey}. CNNs take raw data as their input and extract high-level feature vectors, also known as \textit{embeddings}, using convolutional ﬁlters in multiple layers, followed by some fully connected layers. When dealing with sequential data, such as time-series or multi-view sequences, Recurrent Neural Networks (RNNs) can be used to extract dependencies, as RNN units have dependency connections not only between the subsequent layers, but also into themselves, to learn information from previous inputs \cite{RNNSurvey}. 

The Long Short-Term Memory (LSTM) \cite{LSTM} network is a prominent RNN architecture able to learn a model from both long- and short-term dependencies using learnable gating functions and memory states. LSTM networks are widely used in modern deep learning architectures, and have achieved impressive results on many large-scale machine learning tasks \cite{LSTMOD}. The combination of CNNs and LSTMs has recently been used for several visual recognition tasks, including action recognition \cite{SPL,act3}, face recognition \cite{CSVT, faceLSTM}, facial expression classification \cite{ACII, emotLSTM}, lip reading \cite{lip}, and image captioning and video description \cite{LSTMdesc}. 
A conventional LSTM network can learn a model for the information associated with a single input sequence. However, if two or more dependent sequences of data are simultaneously acquired with some specific dependency, the conventional LSTM networks may not jointly process them inside each LSTM cell, thus not taking advantage of the information present through their dependencies. Instead, the dependent input sequences can thus far only be processed individually by conventional LSTM networks whose results can subsequently be fused, for instance using feature-level or score-level fusion mechanisms \cite{fuse}. Examples of such dependent sequences of data include synchronized audio and visual signals often available in movies \cite{audiovisual}; or horizontal and vertical view sequences with parallax dependencies, as available in multi-view images \cite{CSVT}.

This paper proposes two novel LSTM cell architectures, which tackle the problem above by jointly learning deep models, accepting as inputs two or multiple data sequences that are simultaneously acquired and have some dependency/relationship. The outcome is a jointly learned model that provides richer embeddings to achieve better performance, notably for visual recognition tasks. The proposed cell architectures adopt:
\begin{itemize}
    \item A \textit{Gate-Level Fusion} scheme (GLF-LSTM), modifying the \textit{forget}, \textit{input} and \textit{output} gates of the conventional LSTM architecture 
    to learn from multiple input sequences, providing a fused output for each of these gates. The memory state outputs are controlled by the outcome of the new fused gates, thus providing richer joint information. 
    \item A \textit{State-Level Fusion} scheme (SLF-LSTM), learning the modified \textit{cell} and \textit{hidden} memory states from multiple simultaneous inputs, and then merging the states' outputs to compute the jointly learned embeddings. 
\end{itemize}

To show the efficacy of the novel LSTM cell architectures, they have been used for face recognition with Light Field (LF) images. LF cameras simultaneously capture the intensity of light rays coming from multiple directions in space at a single temporal instant \cite{lensletLF,LF}. An LF image can be rendered to form a multi-view array, offering both intra-view spatial (within each view) and inter-view angular (across views) information, useful for various visual analysis tasks, including biometric recognition \cite{ryrb13,rryb13, rrb16, icip, ear, MLSP, CSVT}, presentation attack detection \cite{TIFS, IETanti, rrb15},  and expression recognition \cite{ACII,ICASSP}. A recent LF-based face recognition method \cite{CSVT} exploits the spatio-angular horizontal and vertical information available in an LF image, by respectively using two independent LSTM recurrent networks whose inputs are VGG-16 embeddings \cite{pvz15}. Although the horizontal and vertical network outputs are fused using a score-level fusion mechanism, this architectural approach cannot fully exploit the relations between the horizontal and vertical views as the horizontal and vertical view sequences are independently processed. Those additional dependencies, notably in terms of parallax, can be further exploited during the learning process by using the proposed LSTM cell architectures, aiming to achieve a better recognition accuracy.

In this paper, each of the novel LSTM cell architectures has been integrated into a deep network for LF-based face recognition, using a Bi-directional LSTM (Bi-LSTM) approach that takes as input ResNet-50 deep embeddings for the horizontal and vertical viewpoint sequences derived from the same LF image. An attention mechanism is also used to selectively focus on the most important jointly learned spatio-angular embeddings, for a more effective learning. The results are two novel face recognition methods, denoted as \textit{ResNet + GLF-LSTM} and \textit{ResNet + SLF-LSTM}, each adopting one of the novel LSTM cell architectures. The new methods have been evaluated on the IST-EURECOM Light Field Face Database (LFFD) \cite{iwbf}, which contains several facial variations, including emotions, actions, poses, illuminations and occlusions, using three challenging evaluation protocols. Results show the superiority of the proposed methods, providing significant face recognition performance improvements regarding the state-of-the-art methods available in the literature.

The rest of the paper is organized as follows: Section II reviews the long short-term memory cell architecture, the basic concepts of LF imaging and the state-of-the-art on LF-based face recognition. The two proposed LSTM cell architectures, notably gate-level fusion and state-level fusion, are presented in Section III. Next, Section IV reports and discusses the performance of the proposed LSTM cell architectures after integration into a deep network for LF-based face recognition. Finally, Section V concludes the paper and proposes some future work.

\section{Background}

This section briefly reviews the conventional LSTM cell architecture, denoted as Conv-LSTM. As the application scenario considered is LF-based face recognition, also the basic concepts of LF imaging and the state-of-the-art on LF-based face recognition are reviewed.

\subsection{Conventional LSTM Cell Architecture}

RNNs can be used to extract higher dimensional dependencies from sequential data \cite{RNNSurvey}. RNN units are called cells, and have connections not only between the subsequent layers, but also into themselves to keep information from previous inputs. The training of a RNN can be done using the back-propagation through time algorithm \cite{backpro}. Traditional RNNs can easily learn short-term dependencies; however, they have difficulties to learn long-term dynamics as the back-propagated gradients can vanish or explode \cite{vanish}. The Long Short-Term Memory (LSTM) is a type of RNN addressing these problems as the LSTM cells allow gradients to also flow unchanged, to avoid the gradient vanishing and exploding problem during training, while learning both long- and short-term dependencies \cite{LSTM,LSTMOD}.

\begin{figure}[!t]
\centering
\includegraphics[width=0.9\columnwidth]{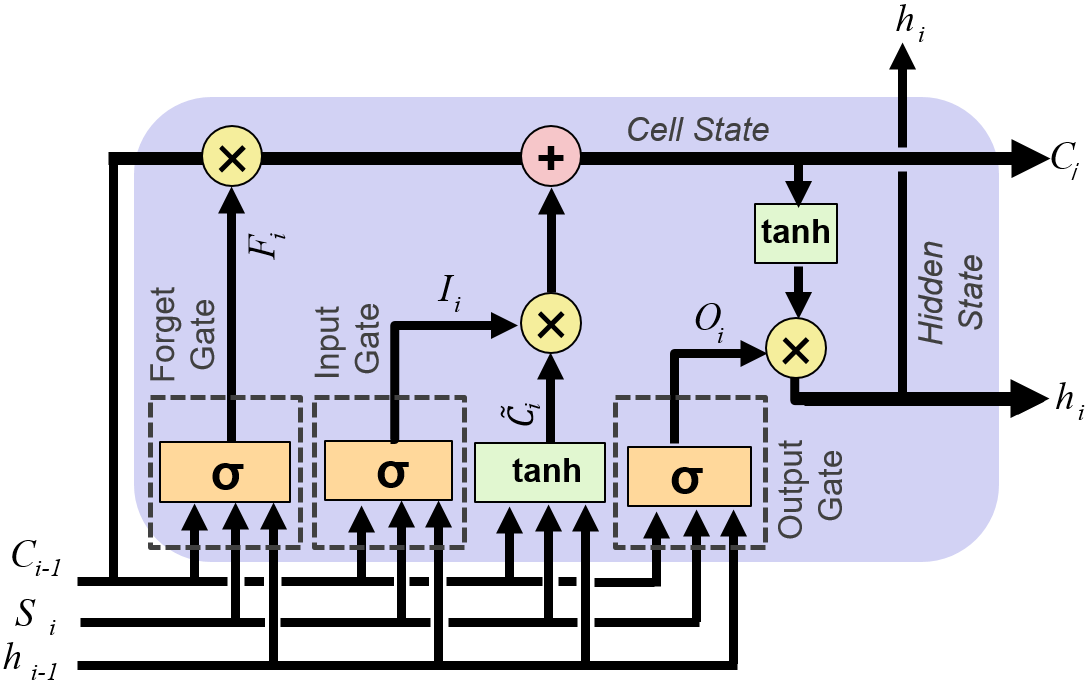}
\caption{Architecture of a conventional LSTM (Conv-LSTM) cell with peephole connections.}
\label{fig:LSTM}
\end{figure}

\begin{figure*}[!t]
\centering
\includegraphics[width=2\columnwidth]{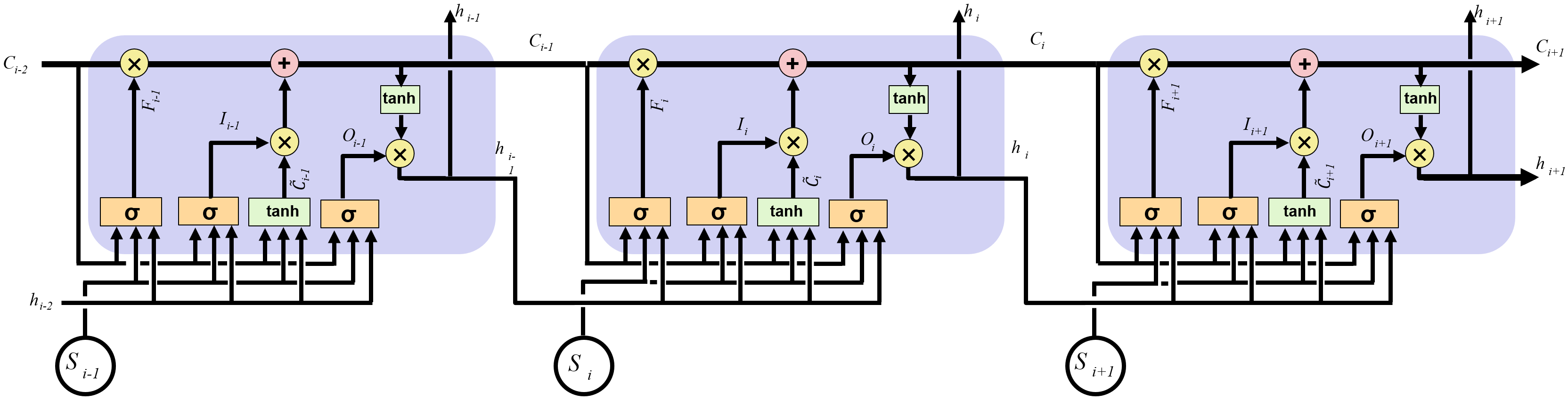}
\caption{An LSTM network composed by several Conv-LSTM cells.}
\label{fig:LSTMNET}
\end{figure*}

Since the introduction of LSTM in 1997 \cite{LSTM}, the conventional LSTM with peephole connections \cite{peep} has been the most commonly used cell architecture for multiple learning tasks \cite{LSTMOD}. An LSTM network is composed by Conv-LSTM cells, whose outputs evolve throughout the network, based on past memory content. The cells have a common cell state, which keeps long-term dependencies along the entire Conv-LSTM cells chain, controlled by two gates, the so-called input and forget gates, thus allowing the network to decide when to forget the previous state or update the current state with new information. The output of each cell, hidden state, is controlled by an output gate taking into account the cell state. Figure \ref{fig:LSTM} illustrates the Conv-LSTM cell architecture with peephole connections, which are connections from the previous cell state to the three gates.

Each of these gates is controlled by a sigmoid activation function, as defined by Equation 1, bounding its output to the [0,1] range:
\begin{equation}
\sigma(x)={(1+{e}^{-x})}^{-1},
\end{equation}

For a sequence component $S_i$, belonging to the input sequence $S$, the input gate, $I_i$, is computed as in Equation (2), based on $S_i$, the previous hidden state $H_{i-1}$, and the previous cell state $C_{i-1}$ (when using peephole connections). The input gate learns how to add information to the cell state. 

\begin{equation}
I_i=\sigma(W_I[S_i+H_{i-1}+C_{i-1}]+b_I),
\end{equation}
where $W_I$ is the input gate weight and $b_I$ is the input gate bias.  

Equation (3) creates a new vector of cell state candidate values, $\tilde{C}_i$, that can be added to the cell state later:
\begin{equation}
\tilde{C}_i=\tanh(W_{\tilde{C}} [H_i+h_{i-1}+C_{i-1}]+b_{\tilde{C}}),
\end{equation}
where $W_{\tilde{C}}$ is the weight of the candidate vector and $b_{\tilde{C}}$ is the bias of the candidate vector. The hyperbolic tangent activation function, $\tanh$, is the non-linearity function used for creating the candidate values, ${\tilde{C}}_i$, defined as:
\begin{equation}
\tanh(x)=2\sigma(2x)-1,
\end{equation}

The forget gate, $F_{i}$, learns how to forget information from the cell state, according to: 
\begin{equation}
F_i=\sigma(W_F[S_i+H_{i-1}+C_{i-1}]+b_F),
\end{equation}
where $W_{F}$ is the forget gate weight and $b_{F}$ is the forget gate bias.

Then, based on $I_{i}$, $F_{i}$, and $\tilde{C}_i$, the previous cell state, $C_{i-1}$, is updated to obtain $C_{i}$ as follows:
\begin{equation}
C_i= F_i \odot C_{i-1} + I_i \odot \tilde{C}_{i},
\end{equation}
where $\odot$ denotes the vector element-wise product. As the output values for $I_{i}$ and $F_{i}$ lie in the range [0,1], Conv-LSTM selectively learns to consider or forget the current input and the previous state.

The current cell state, $C_i$, can then be used for predicting the current cell’s hidden state, $h_i$, according to Equations (7) and (8), thus allowing Conv-LSTM to learn how much from the cell memory should be included into the hidden state:
\begin{equation}
O_i= \sigma(W_O[S_i+H_{i-1}+C_{i-1}]+b_O),
\end{equation}
\begin{equation}
h_i= O_i \odot \tanh(C_i),
\end{equation}
where $W_O$  is the output gate weight and $b_O$ is the output gate bias. The hidden state, $h_i$, is the cell output for the input sequence $S-i$. The hidden state is taken as input by the next Conv-LSTM cell in the LSTM network architecture, as illustrated in Figure \ref{fig:LSTMNET}. This network architecture, also known as uni-directional LSTM network, hereafter only referred as LSTM network, only considers forward relations between LSTM cells. The Bi-directional LSTM (Bi-LSTM) network architecture \cite{bilstm} can also be adopted to exploit both forward (left to right) and backward (right to left) relations within the input sequence. The number of cells in an LSTM network equals the number of inputs, e.g. images/embeddings in the input sequence. The output of each Conv-LSTM cell corresponds to an embedding produced by taking into account the short- and long-term dependencies observed up to that cell’s input image/description.

\subsection{LF Imaging Basics}

The so-called Plenoptic function \textit{P(x,y,z,t,$\lambda$,$\theta$,$\phi$)}, was proposed in 1991 to model the information carried by the light rays at every point in the 3D space (\textit{x,y,z}), in every possible direction ($\theta$,$\phi$), over any wavelength ($\lambda$), and at any time (\textit{t}) \cite{ab91}. The less complex static 4D LF \cite{LF}, \textit{L(x,y,u,v)}, also known as lumigraph \cite{lumi}, was proposed in 1996, by adopting several simplifications on the plenoptic function and may be described by the intersection points of the light rays with two parallel planes \cite{d14}. 

\begin{figure}[!t]
\centering
\includegraphics[width=1\columnwidth]{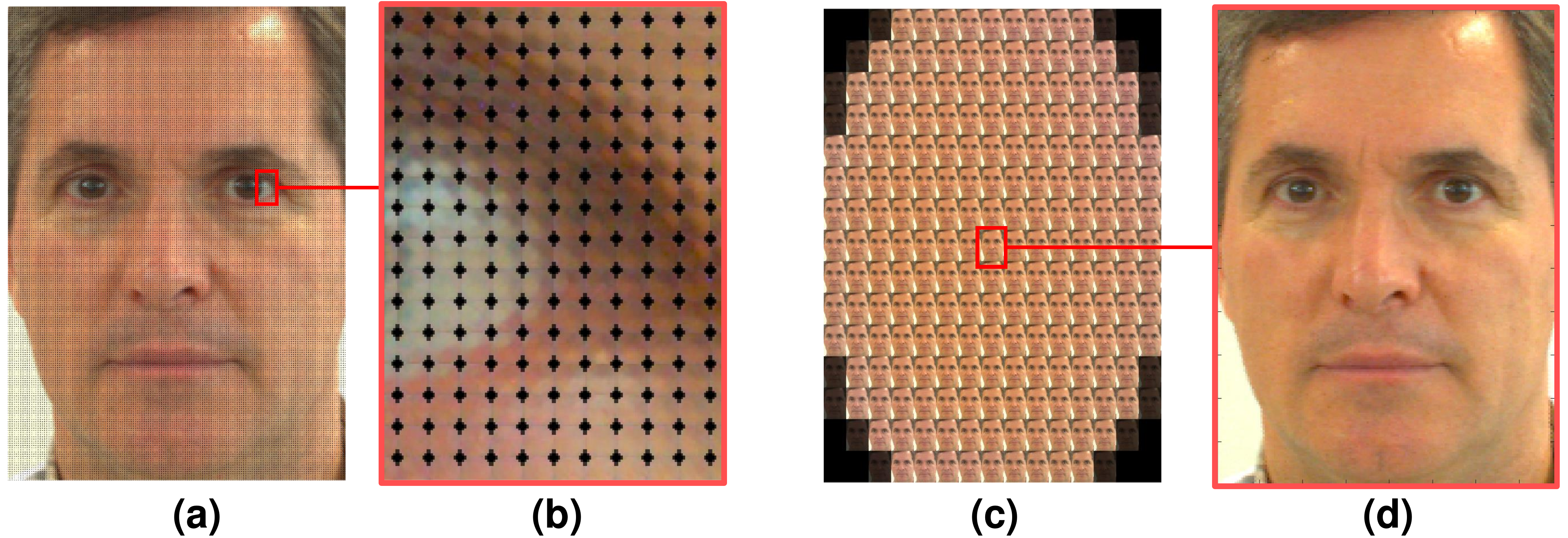}
\caption{LF representation: a) sample micro-images, after colour demosaicing, b) a zoomed detail of the micro-image; c) corresponding multi-view SA array; d) corresponding 2D central SA image.}
\label{fig:LF}
\end{figure}

A lenslet LF camera \cite{lensletLF} includes a digital sensor, main optics and an aperture control similar to regular cameras. The main difference regarding regular cameras comes from placing a micro-lens array at the focal plane of the main lens at a given distance from the sensor. The main lens aims to focus the light rays from the scene object into the microlens array. Then, the micro-lenses split the incoming light cone based on the direction of the incoming rays onto the sensor area of the corresponding micro-lens. A micro-lens array is usually composed by thousands of tiny lenses that are arranged into a rectangular, hexagonal or custom grid. In fact, each micro-lens plays the role of a small camera to acquire a so-called micro-image; Figure \ref{fig:LF}.a shows the full set of micro-images in an LF image, after color demosaicing. The micro-images can then be rendered to extract the so-called Sub-Aperture (SA) images corresponding to different observation viewpoints, forming a multi-view SA array which represents the visual scene. Figure \ref{fig:LF}.b and Figure \ref{fig:LF}.c show the multi-view SA array and the rendered central SA 2D image, respectively, for the LF micro-lens image in Figure \ref{fig:LF}.a. Each 2D SA image corresponds to a slightly different viewpoint of the visual scene, meaning that an LF image captures angular information about the scene, ‘seeing’ it from different angles, which is a distinctive characteristic of this new type of visual sensor.

Since one LF image allows obtaining multiple 2D SA images, two types of information are available for learning: i) the intra-view, spatial information within each 2D view; and ii) the inter-view, angular information between views, associated to the different angular capturing positions of the multiple views. This richer spatio-angular visual representation provides a range of new capabilities to exploit/learn for various visual recognition tasks, notably face recognition \cite{LFsurvey}.

\subsection{LF-based Face Recognition Methods}
A number of face recognition methods exploiting LF information are available in the literature, and are briefly reviewed here. These methods, can be categorized as conventional or deep learning-based, and are summarized in Table \ref{tab:Method}, including information about the publication year, the feature extractor, classifier, and the used datasets.

\subsubsection{Conventional Methods}
There are a few multilinear appearance-based methods able to analyze the high-dimensional LF image information in its native form. It is worth mentioning that none of the multilinear appearance-based methods were originally designed for face recognition. Multilinear Principal Component Analysis (MPCA) \cite{lpv08} is one such method, using tensors for feature extraction, decomposing the LF images which are interpreted as 4D tensors, into a series of multiple projections to capture most of the tensorial input variations. MPCA has been considered for face recognition using LF images in \cite{icip}. Additionally, there are a few methods based on visual descriptors, exploiting the \textit{a posteriori} refocusing capability of LF cameras to improve the quality of the faces in the input image that e.g. might be out-of-focus. In this context, in \cite{ryrb13}, a wavelet energy method selects the best focus image plane whose features are then extracted using the Local Binary Pattern (LBP) descriptor and classified using a Nearest Neighbor (NN) classifier. In \cite{rryb13}, a discrete wavelet transform is used to capture the highest frequency components to create an all-in-focus image, which are then used as input to the LBP descriptor. The recognition of multiple faces available in a single image using an all-in-focus image rendered from an LF image is investigated in \cite{rryb16}, using an LBP descriptor and a Sparse Reconstruction Classifier (SRC) classifier. The exploitation of a set of refocused images rendered from an LF image is studied in  \cite{rrb16}, using different visual descriptors, including LBP, Center-Symmetric LBP (CSLBP), Histogram of Oriented Gradient (HOG), and Binarized Statistical Image Features (BSIF).    
There are also a few visual descriptors exploiting the disparity information available in an LF image. In this context, Light Field LBP (LFLBP) is proposed to exploit the spatio-angular information in an LF images \cite{icip}. Finally, the Light Field Histogram of Gradients (LFHG) descriptor is proposed to consider the orientation and magnitude of LF spatio-angular information \cite{ear} for the recognition task.

\subsubsection{Deep Learning-based Methods}
Recent research on LF-based face recognition has shifted towards deep learning-based methods. The first proposed deep-based method, VGG 2D+Disparity+Depth (VGG-$D^3$) \cite{MLSP}, concatenates embeddings extracted using a fine-tuned VGG network \cite{pvz15} for the LF 2D central SA, disparity, and depth images. More recently, a method combining VGG and Conv-LSTM \cite{CSVT} has been proposed exploiting the multi-perspective LF information using VGG and LSTM networks in sequence, providing more discriminative embeddings for the face recognition task.

\begin{table}[!t]
\centering
\caption{Overview of main available LF-based face recognition methods.}
\setlength
\tabcolsep{2pt}
\begin{tabular}{ l| l| l| l| l| l}
\hline
\textbf{Method}            & \textbf{Year} & \textbf{Approach}    & \textbf{Feat. Extractor} & \textbf{Class.} & \textbf{Dataset} \\\hline 
\hline
MPCA Tensor \cite{lpv08}     & 2008 & Conventional                     & MPCA                      & NN                         & N/A     \\ 
LF Face \cite{ryrb13}         & 2013 & Conventional                           & LBP                       & NN       & Private \\ 
Multi-Face LF \cite{rryb13}   & 2013 &  Conventional                          & LBP; LG ﬁlter             & SRC       &  LiFFID  \\
Super Res. LF \cite{rryb16}   & 2013 & Conventional                           & LBP                       & SRC   & LiFFID  \\
Face MF LF \cite{rrb16} & 2016 & Conventional                           & HOG; LBP; BSIF  & SRC               & LiFFID  \\
LFLBP \cite{icip}          & 2017 &  Conventional                           & LFLBP                     & SVM        & LFFD    \\
LFHG \cite{ear}            & 2018 &  Conventional                         & HOG; LFHDG                & SVM      & LFFD    \\ 
VGG-$D^3$ \cite{MLSP}          & 2018 & Deep                   & VGG                       & SVM        &   LFFD    \\
VGG+ LSTM \cite{CSVT}  & 2019 & Deep                   & VGG;  LSTM           & Soft.    &     LFFD    \\ 

    \hline
\end{tabular}
\label{tab:Method}
\end{table}

\section{Novel LSTM Cell Architectures} 

This section proposes two novel LSTM cell architectures exploiting dependencies between multiple, simultaneously acquired, input sequences. LF-based face recognition is adopted here as a target problem to better explain the structure of the novel architectures.

\subsection{Proposal Intuition}

A conventional LSTM (Conv-LSTM) network can learn a model to describe the information coming from one input sequence, such as a set of 2D images or their embeddings. However, if two or more dependent sequences of data are simultaneously acquired, the Conv-LSTM cell architecture can only process these sequences consecutively, not taking benefit of the information carried out by the dependencies. This limitation is removed by the proposed LSTM cell architectures, designed to jointly learn from such sequences, with each LSTM cell simultaneously receiving multiple sequences as input. A more expressive deep model can then be learned by simultaneously processing all the input sequences, creating richer embeddings for the recognition task. In summary, this paper proposes two novel LSTM cell architectures able to jointly learn from multiple input sequences, unlike the Conv-LSTM. 

\subsection{Target Application: LF-Based Face Recognition}

\begin{figure}[!t]
\centering
\includegraphics[width=0.7\columnwidth]{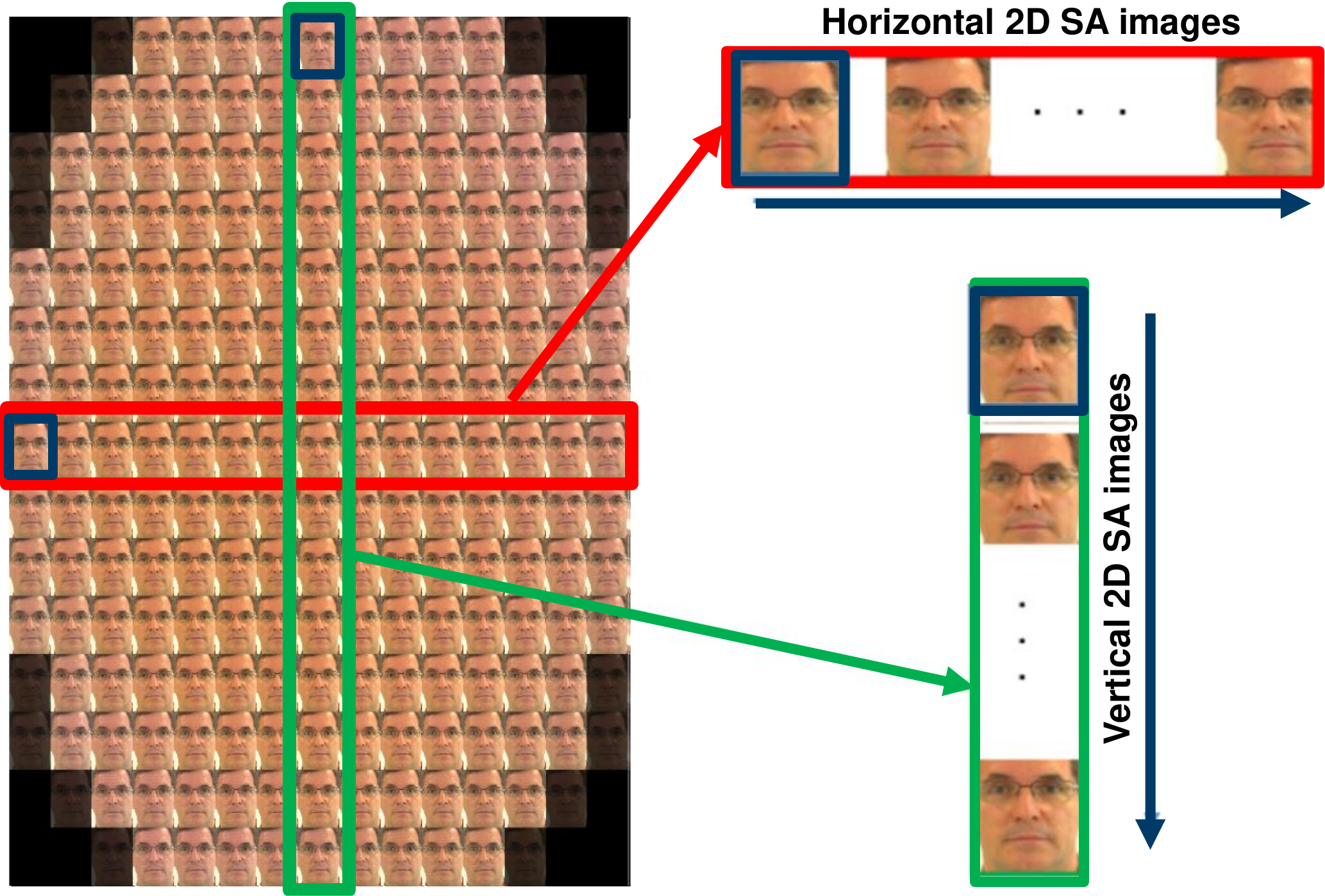}
\caption{Multi-view SA array of face images and the selected set of horizontal and vertical 2D SA images, corresponding to slightly different viewpoints.}
\label{fig:Seq}
\end{figure}

In this paper, LF-based face recognition is adopted as the target application, to demonstrate the added value of the novel LSTM cell architectures. The horizontal and vertical parallaxes (see Figure \ref{fig:Seq}), defined as the displacement or difference in position of an object in two images captured from different perspectives, e.g.,  horizontal or vertical, can represent the viewpoint changes captured in an LF image, defining two sequences of images to be considered as inputs to the LSTM cells. Although the following example considers two input sequences, the same ideas can be used for multiple inputs for different analysis tasks. 

The Conv-LSTM cell architecture can learn the horizontal or the vertical inter-view angular information, as it accepts a single sequence as input. To capture both the horizontal and vertical angular variations, one possibility is to concatenate the LSTM inputs, e.g. a series of features extracted from the horizontal views sequence followed by those extracted from the vertical views sequence. However, the merged sequence may not be the best representation for the dependencies between the horizontal and vertical angular information. 

Since the novel LSTM cell architectures can simultaneously receive multiple input sequences, they can jointly learn the inter-view angular information along the horizontal and vertical directions. As a consequence, an LSTM network built with the novel LSTM cell architectures requires half the number of cells for two input sequences, when compared with the Conv-LSTM network, since each cell now processes two inputs at once. In the considered example, the first novel LSTM cell receives as input the left-most and top SA images, in the middle row and middle column, respectively, as highlighted by the blue boxes in Figure \ref{fig:Seq}. The second LSTM cell processes the second left-most and the second top SA images, and so on. With the new LSTM cell architectures, the relationships within and between horizontal and vertical view sequences can be jointly exploited.

In the proposed LF face recognition methods, SA images are not directly processed by the LSTM network. In fact, it is common practice \cite{LSTMdesc} to first extract spatial embeddings from each input SA image, to first learn from the intra-view spatial information, and then use the LSTM network to also learn the inter-view angular information from the spatial embedding input sequences. In the present proposal, a single LF image is rendered into a multi-view SA array and then horizontal and vertical view sequences are created using selected sets of SA images from the multi-view SA array. A deep CNN is used to extract intra-view spatial embeddings, which are then passed to the novel LSTM architectures. 

The proposed LSTM cell architectures, considering gate-level fusion and state-level fusion schemes, are described in the following. They are illustrated considering two input sequences, named horizontal, $H_i$, and vertical, $V_i$, spatial embeddings computed from the input SA images illustrated in Figure \ref{fig:Seq}. Naturally, the novel LSTM cell architectures can consider any number of correlated input sequences, possibly of different types, addressing other analysis tasks.

\subsection{Gate-Level Fusion LSTM (GLF-LSTM) Cell Architecture}
The first proposed LSTM cell architecture adopts a \textit{gate-level fusion} scheme, learning the horizontal and vertical forget, input and output gates and then merging the horizontal and vertical gates’ outputs to compute the fused gates’ output. In this context, the cell and hidden state outputs are controlled by the fused gates, providing richer joint information to learn a model from the LF angular information.

\begin{figure}[!t]
\centering
\includegraphics[width=0.9\columnwidth]{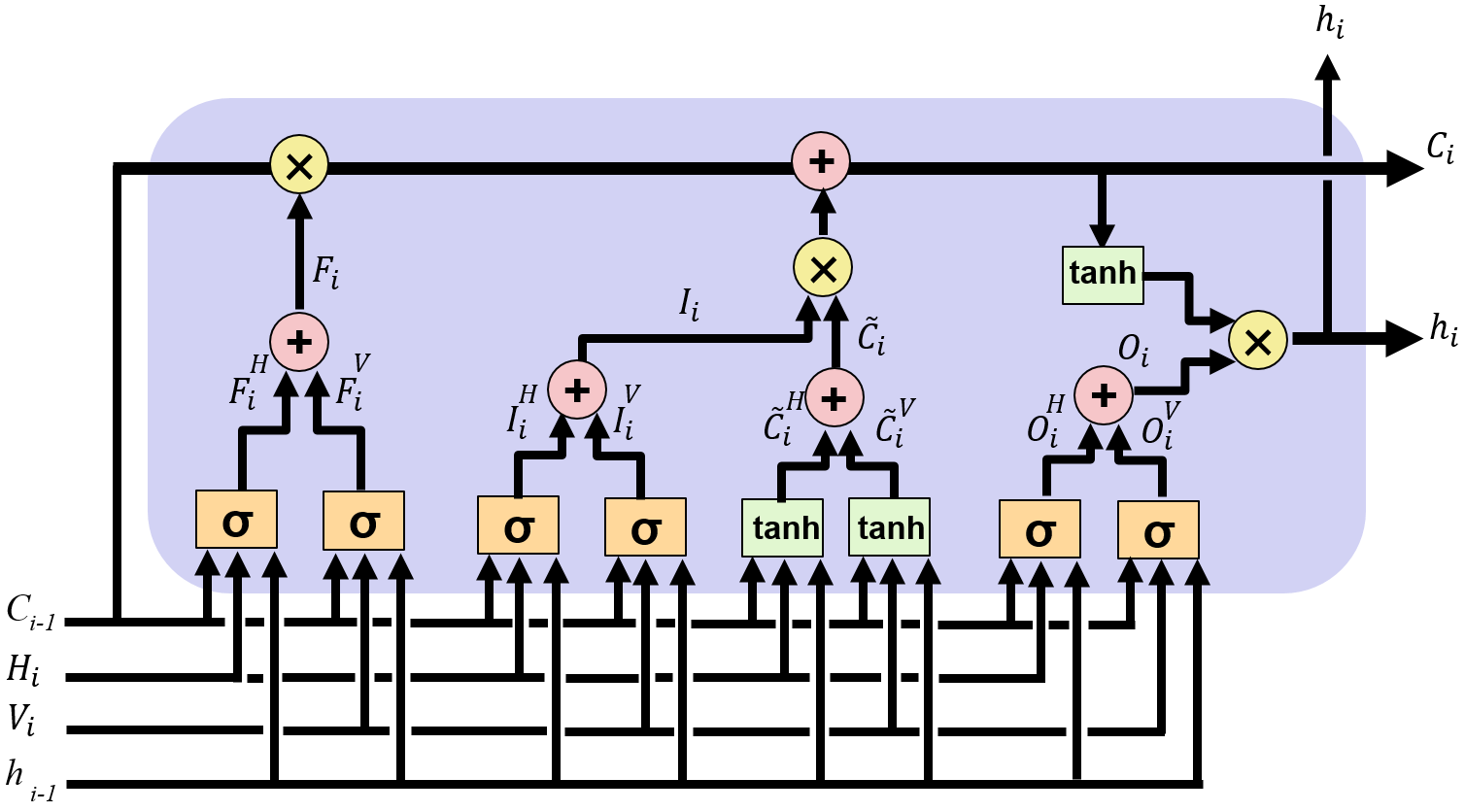}
\caption{Proposed GLF-LSTM cell architecture.}
\label{fig:GLF}
\end{figure}

As illustrated in Figure \ref{fig:GLF}, the horizontal, $I^H_i$, and vertical, $I^V_i$, input gates are computed according to Equations (9) and (10), respectively, independently learning how to add horizontal and vertical information to the cell state. The horizontal and vertical input gates are computed for view number $i$, based on the spatial embeddings $H_{i}$ and $V_{i}$, respectively extracted from the horizontal and vertical multi-view embedding sequences, the previous hidden state $h_{i-1}$, and the previous cell state $C_{i-1}$ as:
\begin{equation}
I^H_i=\sigma(W^H_{I} [H_i+h_{i-1}+C_{i-1}]+b^H_{I}),
\end{equation}
\begin{equation}
I^V_i=\sigma(W^V_{I} [V_i+h_{i-1}+C_{i-1}]+b^V_{I}),
\end{equation}
where $W^H_{I}$ and $W^V_{V}$ are the horizontal and vertical input gates weights, respectively, and $b^H_{I}$ and $b^V_{I}$ are the horizontal and vertical input gates bias, respectively. 

Then, the fused input gate, $I_{i}$, is computed by adding the horizontal and vertical input gates together: 
\begin{equation}
I_i= [I^H_i + I^V_i],
\end{equation}

The horizontal, $\tilde{C}^H_i$, and vertical $\tilde{C}^V_i$, vectors of candidate values are computed according to Equations (12) and (13), respectively. These vectors hold possible horizontal and vertical weights that could be fully/partly added to the fused cell state later. 
Horizontal and vertical candidate vectors are then fused to compute the fused candidate vector, $\tilde{C}_i$, (Equation (14)): 
\begin{equation}
\tilde{C}^H_i=\tanh(W^H_{\tilde{C}} [H_i+h_{i-1}+C_{i-1}]+b^H_{\tilde{C}}),
\end{equation}
\begin{equation}
\tilde{C}^V_i=\tanh(W^V_{\tilde{C}} [V_i+h_{i-1}+C_{i-1}]+b^H_{\tilde{C}}),
\end{equation}
\begin{equation}
\tilde{C}_i= [\tilde{C}^H_i + \tilde{C}^V_i],
\end{equation}
where $W^H_{\tilde{C}}$ and $W^H_{\tilde{V}}$ are the horizontal and vertical weights of the candidate vector, respectively, and $b^H_{\tilde{C}}$ and $b^V_{\tilde{C}}$  are the horizontal and vertical biases of the candidate vector, respectively. 

Next, the horizontal, $F^H_i$ , and vertical, $F^V_i$, forget gates are computed according to Equations (15) and (16), respectively, learning how to forget horizontal and vertical information from the cell state, and after fused to compute the fused hidden state, $F_i$, (Equation (17)). 
\begin{equation}
F^H_i=\sigma(W^H_{F} [H_i+h_{i-1}+C_{i-1}]+b^H_{F}),
\end{equation}
\begin{equation}
F^V_i=\sigma(W^V_{F} [V_i+h_{i-1}+C_{i-1}]+b^V_{F}),
\end{equation}
\begin{equation}
F_i= [F^H_i + F^V_i],
\end{equation}
where $W^H_{F}$ and $W^V_{F}$ are the horizontal and vertical forget gates weights, respectively, and $b^H_{F}$ and $b^V_{F}$ are the horizontal and vertical forget gates biases, respectively. 

Then, based on $I_i$, $F_i$, and $\tilde{C}_i$, the previous cell state, $C_{i-1}$, is updated to obtain $C_{i}$, according to Equations (18), to update the long-term memory observed so far with respect to the horizontal and vertical information observed in the current horizontal and vertical view embeddings:
\begin{equation}
C_i= F_i \odot C_{i-1} + I_i \odot \tilde{C}_{i},
\end{equation}

The horizontal, $O^H_i$ , and vertical, $O^V_i$, output gates are computed according to Equations (19) and (20), respectively, learning how to update hidden state. These gates are then added to compute the fused output gate, $O_i$, according to Equation (21). 

\begin{equation}
O^H_i=\sigma(W^H_{O} [H_i+h_{i-1}+C_{i-1}]+b^H_{O}),
\end{equation}
\begin{equation}
O^V_i=\sigma(W^V_{O} [V_i+h_{i-1}+C_{i-1}]+b^H_{O}),
\end{equation}
\begin{equation}
F_i= [O^H_i + O^V_i],
\end{equation}
where $W^H_{O}$ and $W^V_{O}$ are the horizontal and vertical output gates weights, respectively, and $b^H_{O}$ and $b^V_{O}$ are the horizontal and vertical input gates biases, respectively.

The current cell state, $C_i$, already updated with respect to the jointly learned fused gates and the fused output gate, $F_i$, can then be used for predicting the current cell’s hidden state, $h_i$, according to Equation (22), thus producing the final embedding for the novel GLF-LSTM cell:
\begin{equation}
h_i= O_i \odot \tanh(C_i),
\end{equation}

In our example, the GLF-LSTM cell architecture jointly learns a deep model from LF horizontal and vertical information, in the form of fused gates. It is composed by independent horizontal and vertical gates, so that the computation of the horizontal and vertical input, forget, and output gates can be performed in parallel when implementing this cell architecture.

\subsection{State-Level Fusion LSTM (SLF-LSTM) Cell Architecture}

The second proposed LSTM cell architecture provides a \textit{state-level fusion} scheme, learning the horizontal and vertical cell and hidden states, and then merging their outputs to compute the fused cell and hidden state outputs. 

\begin{figure}[!t]
\centering
\includegraphics[width=0.9\columnwidth]{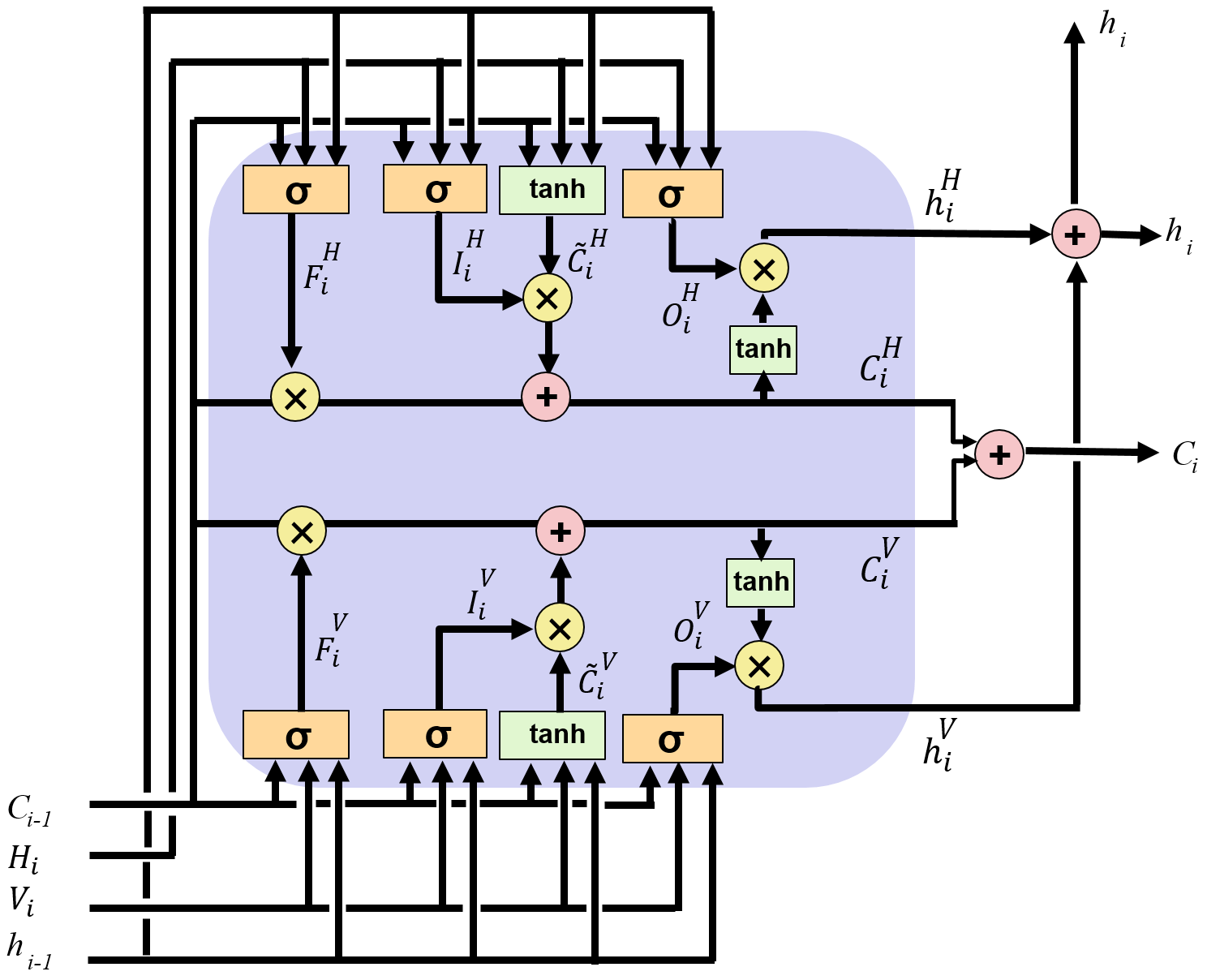}
\caption{Proposed SLF-LSTM cell architecture.}
\label{fig:SLF}
\end{figure}

As illustrated in Figure \ref{fig:SLF}, first, the horizontal elements, including horizontal input gate (Equation (23)), candidate vector (Equation (24)), and forget gate (Equation (25)), are computed based on the horizontal view embedding $H_{i}$, the previous hidden state $h_{i-1}$, and the previous cell state $C_{i-1}$:
\begin{equation}
I^H_i=\sigma(W^H_{I} [H_i+h_{i-1}+C_{i-1}]+b^H_{I}),
\end{equation}
\begin{equation}
\tilde{C}^H_i=\tanh(W^H_{\tilde{C}} [H_i+h_{i-1}+C_{i-1}]+b^H_{\tilde{C}}),
\end{equation}
\begin{equation}
F^H_i=\sigma(W^H_{F} [H_i+h_{i-1}+C_{i-1}]+b^H_{F}),
\end{equation}

Then, based on $I^H_i$, $F^H_i$, and $\tilde{C}^H_i$, the previous cell state, $C_{i-1}$, is updated to obtain the horizontal cell state $C^H_{i}$, according to Equation (26). This means the long-term memory, including the horizontal and vertical information observed so far, is updated with respect to the horizontal information observed in the current view embedding as follows:
\begin{equation}
C^H_i= F^H_i \odot C_{i-1} + I^H_i \odot \tilde{C}^H_{i},
\end{equation}

To learn how to update the horizontal hidden state, the horizontal output gate, $O^H_i$, is computed as:
\begin{equation}
O^H_i=\sigma(W^H_{O} [H_i+h_{i-1}+C_{i-1}]+b^H_{O}),
\end{equation}

The horizontal cell state, $C^H_i$, already updated with respect to the horizontal gates, along with the horizontal output gate, $O^H_i$, can be used for predicting the horizontal cell hidden state, $h^H_i$, according to Equation (28), thus producing the horizontal embedding for the novel SLF-LSTM cell:
\begin{equation}
h^H_i= O^H_i \odot \tanh(C^H_i),
\end{equation}

Next, the vertical elements, including vertical input gate ($I^v_i$), candidate vector ($\tilde{C}^v_i$), forget gate ($F^v_i$), cell state ($C^v_i$), output gate ($O^v_i$), and hidden state ($h^v_i$) are computed, according to Equations (29), (30), (31), (32), (33), and (34), respectively, based on the vertical spatial embedding $V_i$, the previous hidden state $h_{i-1}$, and the previous cell state $C_{i-1}$:
\begin{equation}
I^V_i=\sigma(W^V_{I} [V_i+h_{i-1}+C_{i-1}]+b^V_{I}),
\end{equation}
\begin{equation}
\tilde{C}^V_i=\tanh(W^V_{\tilde{C}} [V_i+h_{i-1}+C_{i-1}]+b^V_{\tilde{C}}),
\end{equation}
\begin{equation}
F^V_i=\sigma(W^V_{F} [V_i+h_{i-1}+C_{i-1}]+b^V_{F}),
\end{equation}
\begin{equation}
C^V_i= F^V_i \odot C_{i-1} + I^V_i \odot \tilde{C}^V_{i},
\end{equation}
\begin{equation}
O^V_i=\sigma(W^V_{O} [V_i+h_{i-1}+C_{i-1}]+b^V_{O}),
\end{equation}
\begin{equation}
h^V_i= O^V_i \odot \tanh(C^V_i),
\end{equation}

Finally, the cell and hidden state outputs, that were independently computed based on horizontal and vertical information, are added together to compute the fused cell state, $C_i$, and the fused hidden state, $h_i$, according to: 
\begin{equation}
C_i= [C^H_i + C^V_i],
\end{equation}
\begin{equation}
h_i= [h^H_i + h^V_i],
\end{equation}

The SLF-LSTM cell architecture jointly learns a model from the LF horizontal and vertical information in the form of fused states composed by independent horizontal and vertical states. The SLF-LSTM parallelism capability is the same as for GLF-LSTM since all the horizontal and vertical learning weights are independently computed.

\section{Proposed Face Recognition Methods} 
The two proposed LSTM cell architectures have been integrated into a deep learning-based method for face recognition with LF images. This section presents these face recognition methods.

\begin{figure*}[!t]
\centering
\includegraphics[width=0.9\linewidth]{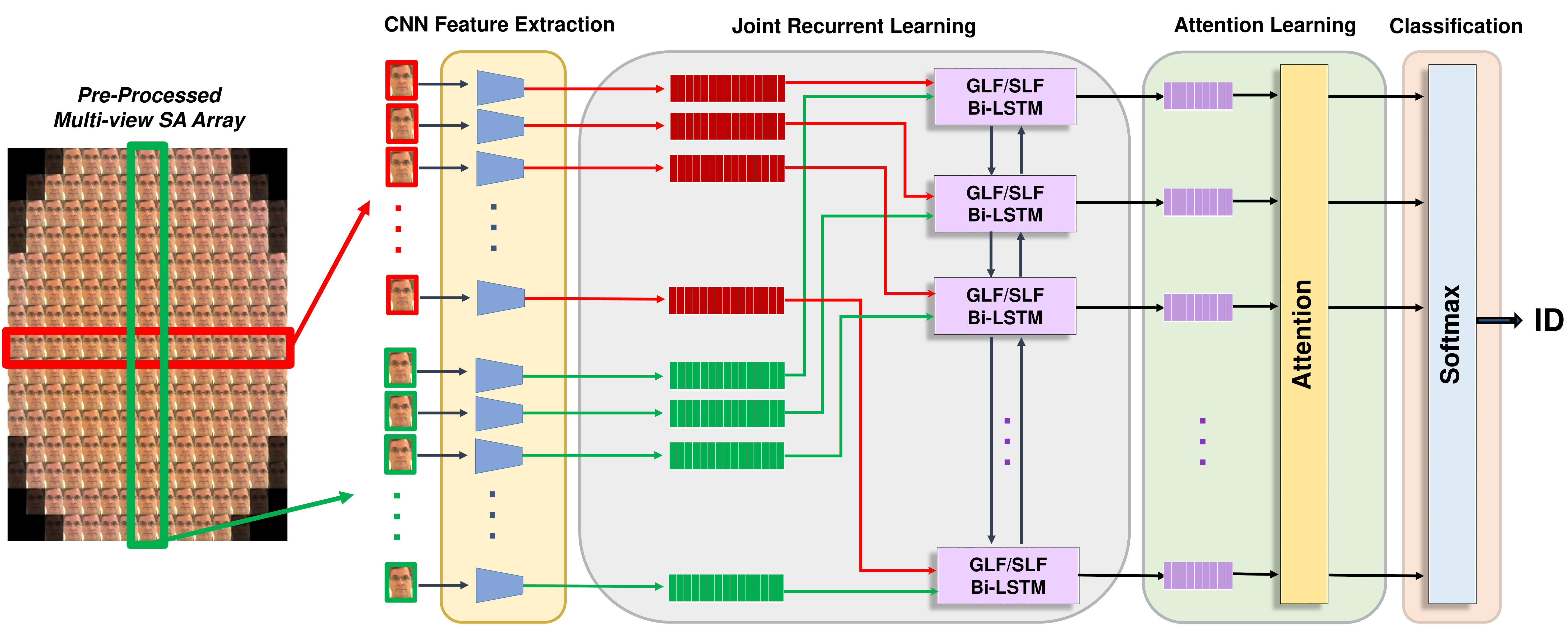}
\caption{Architecture of the proposed face recognition methods adopting the proposed LSTM cell architectures.}
\label{fig:arch}
\end{figure*}

\subsection{Technical Novelty}
The two proposed LSTM cell architectures have been integrated (naturally, one at a time) into a deep network for face recognition. The inputs to the LSTM networks come from a ResNet-50 CNN \cite{resnet} applied to a selected set of horizontal and vertical 2D SA image sequences derived from a single LF image. The differences between the proposed methods and the one described in \cite{CSVT} which adopts Conv-LSTM cells, are: \textit{i}) the two Conv-LSTM networks to be fused at feature-level in \cite{CSVT} are replaced by the new LSTM cell networks proposed in this paper; \textit{ii}) the VGG-16 CNN architecture in \cite{CSVT} is replaced by a more powerful CNN, ResNet-50; \textit{iii}) the LSTM network architecture in this paper is bi-directional as opposed to the uni-directional approach used in \cite{CSVT}; and \textit{iv}) an additional attention layer is added after the recurrent layer, to focus on the most important spatio-angular embeddings.

\subsection{Architecture and Modules}
The architecture of the proposed network for face recognition is presented in Figure \ref{fig:arch}. It should be noted that the two novel LSTM cell architectures lead to two different recognition methods. The proposed methods are composed of a pre-processing module and four sub-networks, notably CNN feature extraction, joint recurrent learning, attention learning, and classification, whose descriptions are provided in the following:

\subsubsection{Pre-processing}
First, LF raw (LFR) input images are rendered using the Light Field Toolbox v0.4 software \cite{LFTool}, to create the multi-view SA array, which includes $15\times15$ SA images per LF image, each with a spatial resolution of $625\times434$ pixels. Then, the face region within each SA image is cropped and resized to $224\times224$ pixels, which is the expected input size for the ResNet-50 network. The middle row and the middle column SA images are independently organized into two sequences, each including fifteen SA images. 
These images represent viewpoint changes along the horizontal and vertical directions with maximum parallax, thus better capturing the LF information coming from multiple directions. 

\subsubsection{CNN feature extraction} Each selected SA image is fed to a pre-trained ResNet-50 CNN \cite{resnet}, to extract a spatial embedding with a fixed length of 2048 elements. As ResNet-50 has been pre-trained on the large-scale VGG-Face 2 dataset \cite{csxpz18}, the proposed method will not suffer from overfitting and thus there is no additional training performed at this stage. ResNet-50 \cite{resnet} has been selected here as it delivers better results than other CNN architectures in the context of the proposed face recognition systems. A comprehensive analysis of the impact of two other CNN architectures, namely SE-ResNet-50  \cite{seresnet} and VGG-16 \cite{pvz15}, on the performance of the proposed face recognition method, has been performed and results are presented in Section V-D.

\subsubsection{Joint recurrent learning} The extracted horizontal and vertical ResNet-50 spatial embeddings are provided to one of the novel LSTM networks proposed in Section 3. A Bi-LSTM network incorporates the novel LSTM cell architectures to capture both the joint forward and backward relationships within and between the sequences. The Bi-LSTM network architecture can be modeled as two uni-directional LSTM networks, respectively analyzing the input sequences in the forward and backward directions. Finally, the output hidden states of the two uni-directional LSTM networks are concatenated to produce the Bi-LSTM network's outputs.

\subsubsection{Attention learning} An attention mechanism \cite{att4} is used to selectively focus on the most important LSTM embeddings, corresponding to different viewpoints. It can boost the recognition performance by assigning higher weights to the more relevant spatio-angular embeddings while ignoring the spurious perspectives. To this end, the Bi-LSTM outputs are multiplied by a set of trainable parameters. Equations (37) and (38) compute a score \(a_{i}\) to measure the attention level for \(h_{i}\) which is the output hidden state vector for the \(i^{th}\) LSTM cell, corresponding to the \(i^{th}\) viewpoint:
\begin{equation}
u_{i}=tanh(W_{h}h_{i}+b_{h})\;\;\;\; \forall_{i}\in(1,...,n)
\end{equation}
\begin{equation}
a_{i}=\frac{e^{u_{i}}}{\sum_{j=1}^{n}{e^{u_{j}}}}\;\;\;\; \forall_{i}\in(1,...,n)
\end{equation}
where, \textit{n} is the number of viewpoints and \(W_{h}\) and \(b_{h}\) are the trainable weights and biases for the hidden states \(h_{i}\), respectively. Finally, \textit{AttEmb}, the attention layer’s output embedding, is computed as: 
\begin{equation}
{AttEmb=\sum_{i=1}^{n}{a_{i}h_{i}}}
\end{equation}

\subsubsection{Classification} The set of attention layer’s outputs, corresponding to the attentively learned spatio-angular embeddings, are used as input to a softmax classifier. The classification sub-network uses a softmax activation function that squashes the input values into an output vector in which each element takes a value in the range of $[0,1]$. This sub-network then quantifies the agreement between the squashed vector and the labels to perform face recognition.	

\subsection{Implementation and Training Details}\label{sec:implementation}
The optimal parameters values capable of achieving the best results are summarized in Table \ref{tab: parameter}. This table includes the values for CNN feature extraction, joint recurrent learning, and attention learning sub-networks along with for the whole deep network. 
The entire architecture has been implemented using TensorFlow \cite{tensorflow} with Keras backend \cite{keras} and trained using four Nvidia RTX 2080 Ti GPUs.

\begin{table}[!t]
\centering
\caption{Best parameter values obtained for the proposed face recognition methods.}
\setlength
\tabcolsep{4pt}
\begin{tabular}{ l| l| l}
\hline
\textbf{Sub-Network} & \textbf{Parameter} & \textbf{Setting} \\
\hline
\hline
    CNN Feature Extraction & Architecture  & ResNet-50  \\
      & Pre-trained Model  & VGG-Face2  \\
      & Number of Inputs  & $15 \times 2$  \\
      & Embedding Layer  & Average Pooling  \\
      & Embedding Size  & 2048  \\
    \hline

    Joint Recurrent Learning & Cell Architecture  & GLF/SLF LSTM  \\
      & Number of Inputs  & $15 \times 2$  \\
      & Number of Outputs  & 15 \\
      & Hidden Layer Size  & $256 \times 2$  \\
      & Dropout Rate  & $0.1$  \\
      & Network Architecture  & Bi-directional  \\
    \hline

    Attention Learning & Activation Function  & Softmax  \\
    \hline    

     Full Network  & Mini-batch Size & 100 \\
      & Loss Function & Cross-entropy \\
     & Optimizer & rmsprop \\
     & Metric & Accuracy \\
     & Number of Epochs & 1000\\
    \hline
\end{tabular}
\label{tab: parameter}
\end{table}

\section{Experiments and Performance Assessment} 
This section presents the dataset and test protocols used, the state-of-the-art methods considered for benchmarking, and the obtained performance results along with their analysis. It also includes ablation experiments to investigate the impact of each individual sub-network of the proposed deep networks on the overall face recognition performance.

\subsection{Dataset}
\begin{figure}[!t]
\centering
\includegraphics[width=1\columnwidth]{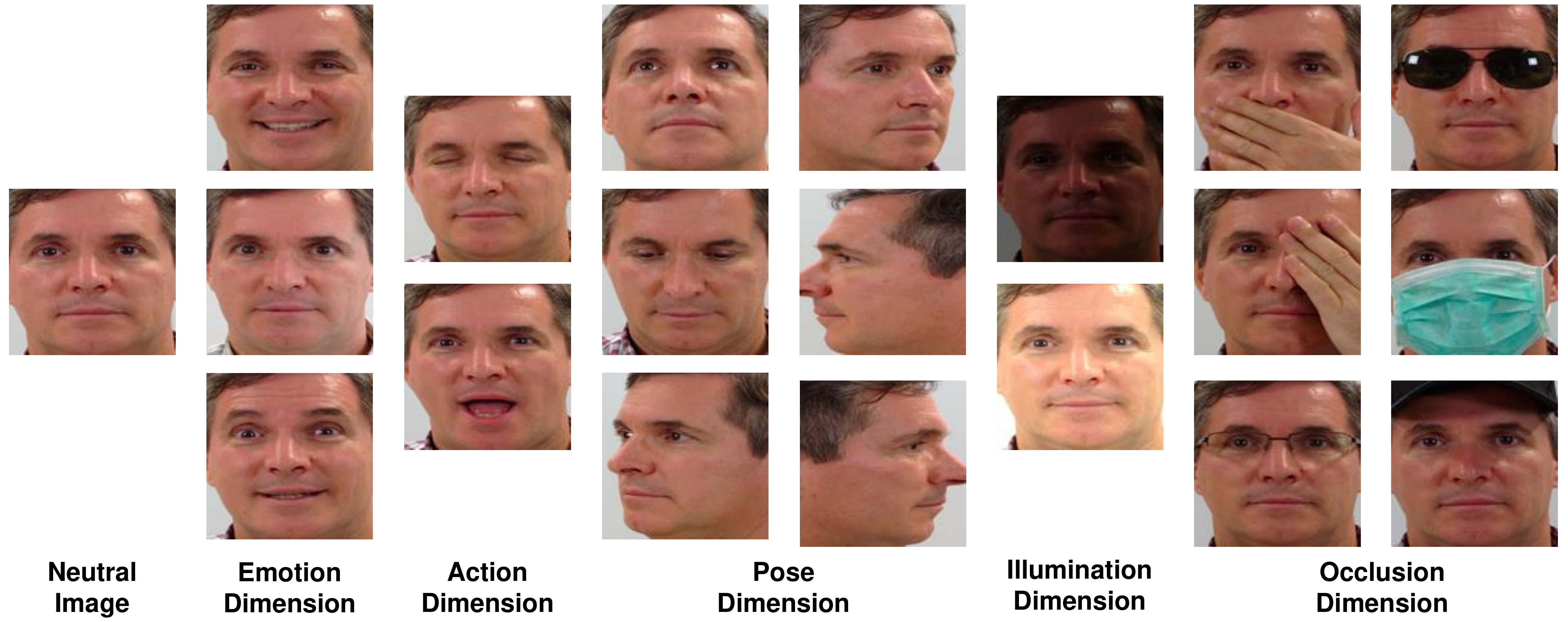}
\caption{IST EURECOM LFFD: Illustration of the set of cropped central SA image variations for a single subject.}
\label{fig:LFDB}
\end{figure}

The IST-EURECOM Light Field Face Database (LFFD) \cite{iwbf} consists of LF face images captured by a Lytro ILLUM camera \cite{Lytro} and will be used here for performance assessment purposes. The IST-EURECOM LFFD includes 4000 LF images, captured from 100 subjects, in two separate acquisition sessions with a temporal separation between 1 and 6 months. Each session contains 20 LF images per subject with different facial variations including facial emotions, actions, poses, illuminations, and occlusions, as illustrated in Figure \ref{fig:LFDB}.  

\begin{figure*}[!t]
\centering
\includegraphics[width=1\linewidth]{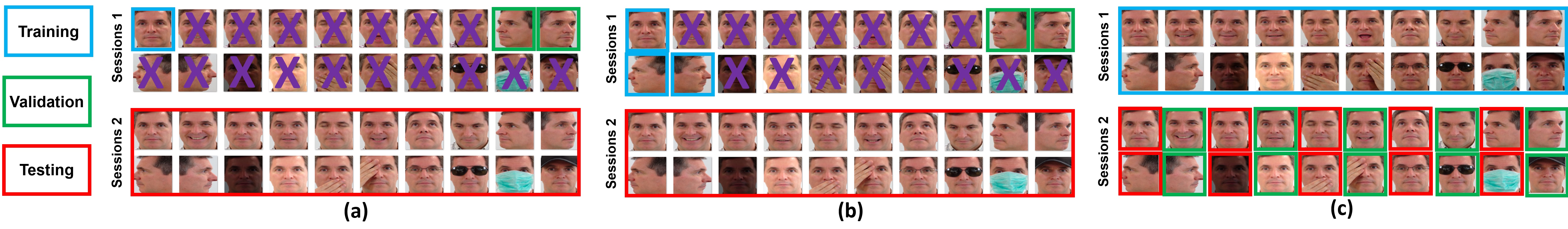}
\caption{IST-EURECOM LFFD division into training, validation and testing sets (a) Protocol 1; (b) Protocol 2; and (c) Protocol 3.}
\label{fig:Prot}
\end{figure*}

\subsection{Test Protocols}
This paper proposes three test protocols with practical meaningfulness to assess the performance of the novel LSTM cell architectures in the context of the face recognition method described in Section 4. The test protocols are defined as follows:

\subsubsection{Protocol 1} The training set contains only the neutral LF images from the first acquisition session (one image per subject), while the validation set includes the left and right half-profile images from the same acquisition session (two images per subject), thus corresponding to a low-complexity enrolment scenario. Testing is done by considering all facial variations captured in the second acquisition session as illustrated in Figure \ref{fig:Prot}.a. This scenario assumes that recognition should be robust to real-life conditions where face images may be captured at different times and in less constrained conditions, for instance with facial expressions or partial occlusions. This protocol represents the \textit{one shot learning} \cite{oneshot} problem that has been considered as one of the most difficult challenges in face recognition \cite{oneshotface}.

\subsubsection{Protocol 2} The training set contains the neutral plus the left and right full-profile LF images from the first acquisition session (three images per subject). The validation set, on the other hand, includes the left and right half-profile images from the same acquisition session (two images per subject) and the test set includes all the LF images from the second acquisition session, as illustrated in Figure \ref{fig:Prot}.b. This protocol also assumes a rather simple and quick enrolment phase, corresponding to a low-complexity enrolment scenario while being slightly less challenging than the first protocol in terms of recognition performance. 

\subsubsection{Protocol 3} The training set contains all twenty dataset face variations captured during the first acquisition session, while the validation and testing sets each consider half of the second session's images, as illustrated in Figure \ref{fig:Prot}.c. 
This scenario assumes a more complex acquisition phase, considering more training images, under the assumption that the increased complexity will result in better training, thus a better recognition performance.

The three protocols correspond to cooperative user scenarios, offering different trade-offs in terms of the initial enrolment and training complexity versus the expected recognition performance. The first and second protocols have multiple practical applications, such as access control systems, where the users can be registered/enrolled into the system by quickly taking a mugshot, including a frontal-view and side-view photos in a controlled setup. Then, the goal is to recognize a person from an image captured at a different time, and possibly in non-ideal conditions, e.g. exhibiting unpredictable facial variations. The third protocol corresponds to a more cooperative user scenario, targeting applications with increased security requirements, where the users are willing to cooperate more during the registration phase, simulating different facial variations, over a range of expressions, actions, poses, illuminations, and occlusions, to capture as many variations as possible during the enrollment phase, so that the recognition system can more effectively recognize users during the daily system operation.

\subsection{Benchmarking Face Recognition Methods}
The LF-based face recognition methods considered for benchmarking purposes include Depth Local Binary Patterns (DLBP \cite{DLBP}, Multi-linear Principle Component Analysis (MPCA) \cite{lpv08}, LFLBP \cite{icip}, HOG+HDG \cite{ear}, VGG 2D+Disparity+Depth (VGG-$D^3$) \cite{MLSP}, and VGG+conv-LSTM \cite{CSVT}. These methods are discussed in Section II-C. It should be noted that all the tested recognition methods were re-implemented by the authors of this paper since this was required for testing with the novel proposed protocols. Performance results were obtained considering the best parameter settings for each method reported in the respective original papers. Additionally, Spatio-temporal LSTM (ST-LSTM) \cite{LSTMact2} is a multi-input LSTM cell architecture available in the literature, allowing the fusion of two types of sequences, can be used for LF-based face recognition. In the original reference, \cite{LSTMact2}, the ST-LSTM architecture is designed to jointly learn a body skeleton, in the form of joint positions, and visual texture data, for activity recognition. The performance of using this cell architecture for LF face recognition has also been considered in the benchmarking study. When presenting the results, the two newly proposed face recognition methods, GLF-LSTM and SLF-LSTM cell architectures, are labelled as ResNet-50 + GLF-LSTM and ResNet-50 + SLF-LSTM. Rank-1 recognition rates are used to report the obtained results.

\subsection{Comparison between CNN Architectures for Feature Extraction}
Table \ref{tab: CNN} presents the performance of the proposed face recognition methods, when the selected CNN architecture used in feature extraction sub-network, ResNet-50 \cite{resnet}, is replaced by two other CNN architectures, notably SE-ResNet-50 \cite{seresnet} and VGG-16 \cite{pvz15}. It is worth mentioning that all these architectures were pre-trained on the same large-scale VGG-Face 2 dataset \cite{csxpz18}. Results are presented for the Conv-LSTM as well as for the novel GLF-LSTM and SLF-LSTM cell architectures when considering the three test protocols described in Section V-B. The results clearly show the recognition performance benefits of using ResNet-50 \cite{resnet} over the two other architectures, thus justifying its selection for the proposed face recognition methods.

\begin{table}[!t]
\centering
\caption{Comparison of three CNN feature extraction architectures in the context of the proposed methods.}
\setlength
\tabcolsep{1.5pt}
\begin{tabular}{  l| l| l| l| l| l| l}
\hline

\hline
    \textbf{CNN}  & \textbf{CNN} & \textbf{LSTM} & \textbf{Protocol} & \textbf{Protocol} & \textbf{Protocol} & \textbf{Mean}\\ 
    \textbf{Architecture} & \textbf{Model} & \textbf{Type} & \textbf{1} & \textbf{2} & \textbf{3} & \\ 
    \hline\hline
    VGG-16 \cite{pvz15} & VGG-Face2 & Conv &  84.80 &  90.80  &  95.70  & 90.43  \\
    SE-ResNet-50 \cite{seresnet} & VGG-Face2& Conv &  96.30  &  96.95  &  98.80 &  97.35 \\
    ResNet-50 \cite{resnet} & VGG-Face2& Conv &  \textbf{97.10}  &  \textbf{97.40}  &  \textbf{99.10}  & \textbf{97.86}  \\
    \hline
    VGG-16 \cite{pvz15}  & VGG-Face2 &  GLF & 86.75  &  92.10  &  96.40  & 91.75  \\
    SE-ResNet-50 \cite{seresnet} & VGG-Face2&  GLF &  97.00  &  97.55  &  99.10  & 97.88  \\
    ResNet-50 \cite{resnet} & VGG-Face2&  GLF &  \textbf{97.75}  & \textbf{98.00}  & \textbf{99.80}   & \textbf{98.51}  \\
    \hline
    VGG-16 \cite{pvz15} & VGG-Face2 &  SLF & 86.95  &  92.60  &  97.00  & 92.18 \\
    SE-ResNet-50 \cite{seresnet} & VGG-Face2&  SLF & 97.00   &  97.30  & 99.10   &  97.80 \\
    ResNet-50 \cite{resnet} & VGG-Face2&  SLF &  \textbf{97.65}  & \textbf{98.00}  & \textbf{99.60}   & \textbf{98.41}  \\
    
    \hline
\end{tabular}
\label{tab: CNN}
\end{table}

\subsection{Comparison to other Fusion Strategies and Joint LSTM Architectures}
Table \ref{tab: fusion} presents the face recognition performance when different strategies are used to fuse the information extracted from the horizontal and vertical LF view sequences. The first and second rows in Table \ref{tab: fusion}, respectively, show the individual performance results obtained by considering the horizontal and vertical sequences separately, i.e., when no fusion strategy has been used. The third row in Table \ref{tab: fusion}, denoted by \textit{Feature-Level Fusion 1}, presents the results when concatenating the horizontal and vertical spatial embeddings together, so that the size of the input sequence to the LSTM network is $15\times4096$. Alternatively, the forth row in Table \ref{tab: fusion} presents the results of \textit{Feature-Level Fusion 2} strategy by first processing the horizontal images followed by the vertical ones, thus creating a concatenated sequence with size $30\times2048$ to be used as the LSTM network input. \textit{Score-Level Fusion} (the fifth row of Table \ref{tab: fusion}) employs two independent LSTM networks for the horizontal and vertical views, whose classification scores are finally added together to perform face recognition. 
Apart from fusion strategies, the recognition performance when using the ST-LSTM cell architecture \cite{LSTMact2} for the LF face recognition task is also reported in the sixth row of Table \ref{tab: fusion}. Finally, the performance of the novel GLF-LSTM and SLF-LSTM cell architectures are presented in the the seventh and eighth rows of the table. The performance results clearly show the superiority of novel cell architectures in comparison with the five fusion strategies and another LSTM cell architecture to deal with horizontal and vertical LF view sequences.

\begin{table}[!t]
\centering
\caption{Comparison of the novel LSTM cell architectures with five alternative fusion strategies and one available LSTM cell architecture for LF-based face recognition.}
\setlength
\tabcolsep{4pt}
\begin{tabular}{  l| l| l| l| l}
\hline
    \textbf{Fusion  Strategy} & \textbf{Protocol 1} & \textbf{Protocol 2} & \textbf{Protocol 3} & \textbf{Mean}\\ 
    \hline\hline
    No Fusion (Hor. Views)& 96.90  &  97.25  &  99.10  &  97.75 \\
    No Fusion (Ver. Views) & 96.45  &  97.30  &  99.20  & 97.65  \\
    Feature-Level Fusion 1 & 97.00  &  97.40  &  99.20  &  97.86 \\
    Feature-Level Fusion 2 &  97.10  &  97.40  &  99.10  & 97.86  \\
    Score-Level Fusion  &  97.05  &  97.45  & 99.20   &  97.90 \\
    ST-LSTM  \cite{LSTMact2}  &  97.25  & 97.65 & 99.30 &  98.06 \\
    GLF-LSTM  &    \textbf{97.75}  & \textbf{98.00}  & \textbf{99.80}   & \textbf{98.51}\\
    SLF-LSTM  &    97.65  & \textbf{98.00}  & 99.60   & 98.41   \\

    \hline
\end{tabular}
\label{tab: fusion}
\end{table}

\subsection{Comparison to State-of-the-art Recognition Methods}

\begin{table*}[]
\centering
\caption{Protocol 1 assessment: Face recognition rank-1 for the proposed and benchmarking recognition methods.}
\setlength
\tabcolsep{10pt}
\begin{tabular}{l|l|l|l|l|l|l|l}
\hline
Method          & \textbf{Year} & \textbf{Emot.} & \textbf{Action}  & \textbf{Pose}    & \textbf{Illum.} & \textbf{Occ.} & \textbf{Mean}    \\
\hline
\hline

DLBP \cite{DLBP}            & 2014 & 59.25             & 64.50 & 30.33 & 24.50      & 22.33   & 36.55 \\
MPCA \cite{lpv08}          & 2017 & 36.75             & 33.50 & 07.50  & 14.50     & 19.67   & 20.30 \\
LFLBP \cite{icip}          & 2017 & 34.25             & 31.00 & 10.17 & 17.00      & 13.17   & 18.65 \\
HOG+HDG \cite{ear}         & 2018 & 62.25             & 62.50 & 12.00 & 62.00      & 41.33   & 40.90 \\
VGG-$D^3$ \cite{MLSP}         & 2018 & 99.50             & 99.00 & 56.50 & 99.00      & 75.50   & 79.50 \\
VGG-16 + Conv-LSTM \cite{CSVT} & 2020 & 99.25             & 99.50 & 71.67 & 99.00      & 91.17   & 88.55\\
\textbf{ResNet-50 + GLF-LSTM} & 2020  & \textbf{100.00} & \textbf{100.00}  & \textbf{94.50}  &  \textbf{100.00} & 98.00 & \textbf{97.75} \\
\textbf{ResNet-50 + SLF-LSTM} & 2020  & \textbf{100.00} & \textbf{100.00 } & 93.83  &  \textbf{100.00} & \textbf{98.33} & 97.65 \\
\hline
\end{tabular}
\label{tab: Res1}
\end{table*}

\begin{table*}[]
\centering
\caption{Protocol 2 assessment: Face recognition rank-1 for the proposed and benchmarking recognition methods.}
\setlength
\tabcolsep{10pt}
\begin{tabular}{l|l|l|l|l|l|l|l}
\hline
Method          & \textbf{Year} & \textbf{Emot.} & \textbf{Action}  & \textbf{Pose}    & \textbf{Illum.} & \textbf{Occ.} & \textbf{Mean}    \\

\hline
\hline
DLBP \cite{DLBP}           & 2014 & 89.50             & 89.00 & 72.50 & 65.00      & 63.33   & 73.60 \\
MPCA \cite{lpv08}           & 2017 & 68.50             & 68.50 & 20.50 & 32.00      & 41.00   & 42.85 \\
LFLBP \cite{icip}          & 2017 & 67.00             & 70.50 & 38.50 & 46.00      & 55.67   & 53.75 \\
HOG+HDG \cite{ear}         & 2018 & 80.00             & 79.00 & 21.34 & 67.50     & 65.00   & 59.20 \\
VGG-$D^3$ \cite{MLSP}         & 2018 & 97.25             & 93.00 & 86.34 & 96.00      & 72.33   & 85.95 \\
VGG-16 + Conv-LSTM \cite{CSVT} & 2020 & 98.50             & 99.00 & 92.00 & 98.00      & 83.00   & 91.95 \\
\textbf{ResNet-50 + GLF-LSTM} &  2020 &  \textbf{100.00}  & \textbf{100.00}  & 97.83  & \textbf{100.00} & \textbf{95.50} & \textbf{98.00 }\\
\textbf{ResNet-50 + SLF-LSTM} & 2020  &  \textbf{100.00}  & \textbf{100.00}  & \textbf{98.00}  & \textbf{100.00} & 95.33 & \textbf{98.00 }\\
\hline
\end{tabular}
\label{tab: Res2}
\end{table*}

\begin{table*}[]
\centering
\caption{Protocol 3 assessment: Face recognition rank-1 for the proposed and benchmarking recognition methods.}
\setlength
\tabcolsep{10pt}
\begin{tabular}{l|l|l|l|l|l|l|l}
\hline
Method          & \textbf{Year} & \textbf{Emot.} & \textbf{Action}  & \textbf{Pose}    & \textbf{Illum.} & \textbf{Occ.} & \textbf{Mean}    \\

\hline
\hline
DLBP \cite{DLBP}           & 2014 & 56.50             & 64.00 & 69.66 & 75.00      & 69.66   & 63.70 \\
MPCA \cite{lpv08}           & 2017 & 48.00             & 89.00 & 65.00 & 63.00      & 64.66   & 50.30 \\
LFLBP \cite{icip}          & 2017 & 52.50             & 96.00 & 87.66 & 76.00      & 87.66   & 65.80 \\
HOG+HDG \cite{ear}        & 2018 & 61.00             & 93.00 & 83.33 & 80.00      & 83.33   & 67.10 \\
VGG-$D^3$ \cite{MLSP}         & 2018 & 94.00             & 98.00 & 98.00 & 97.00      & 98.33   & 97.40 \\
VGG-16 + Conv-LSTM \cite{CSVT} & 2020 & \textbf{100.00}            & \textbf{100.00}& 96.33 & \textbf{100.00}     & 98.66   & 98.80 \\
\textbf{ResNet-50 + GLF-LSTM} & 2020  & \textbf{100.00} &  \textbf{100.00} & \textbf{100.00}   & \textbf{100.00}  &  \textbf{99.33}  & \textbf{99.80} \\
\textbf{ResNet-50 + SLF-LSTM} & 2020  &    \textbf{100.00} &  \textbf{100.00} & 99.33 &  \textbf{100.00}     & \textbf{99.33} & \textbf{99.60}    \\
\hline
\end{tabular}
\label{tab: Res3}
\end{table*}

Tables \ref{tab: Res1}, \ref{tab: Res2}, and \ref{tab: Res3} report the rank-1 recognition rates obtained for the proposed recognition methods and the 6 selected benchmarking methods, for test Protocols 1, 2, and 3, respectively. These results are presented for the five recognition variations, corresponding to the LFFD dimensions (shown in Figure \ref{fig:LFDB}), notably emotions (including neutral face), actions, poses, illuminations and occlusions, using all the test set images as defined for each test protocol; the best results and the proposed methods are highlighted in bold. 

The obtained rank-1 recognition results demonstrate the superiority of the deep learning-based methods when compared to non-deep learning-based methods, including DLBP \cite{DLBP}, MPCA \cite{lpv08}, LFLBP \cite{icip} and HOG+HDG \cite{ear}. The results also show that the proposed recognition methods adopting the two novel LSTM cell architectures achieve better performance than VGG-$D^3$ \cite{MLSP} as well as the best performing benchmarking method, the Conv-LSTM method \cite{CSVT}, for all the face recognition variations considered. The added value of the proposed recognition methods is more evident for the more challenging test Protocols 1 and 2, where limited amounts of training data are available. Finally, performance results show that the proposed ResNet-50 + GLF-LSTM recognition method works slightly better than the ResNet-50 + SLF-LSTM method. This is mainly due to establishing a learning interaction between both LSTM gates and memory states as compared to the ResNet + SLF-LSTM, which only learns joint memory states.

\subsection{Feature Space Exploration}
\begin{figure*}[!t]
\centering
\includegraphics[width=1.25\columnwidth]{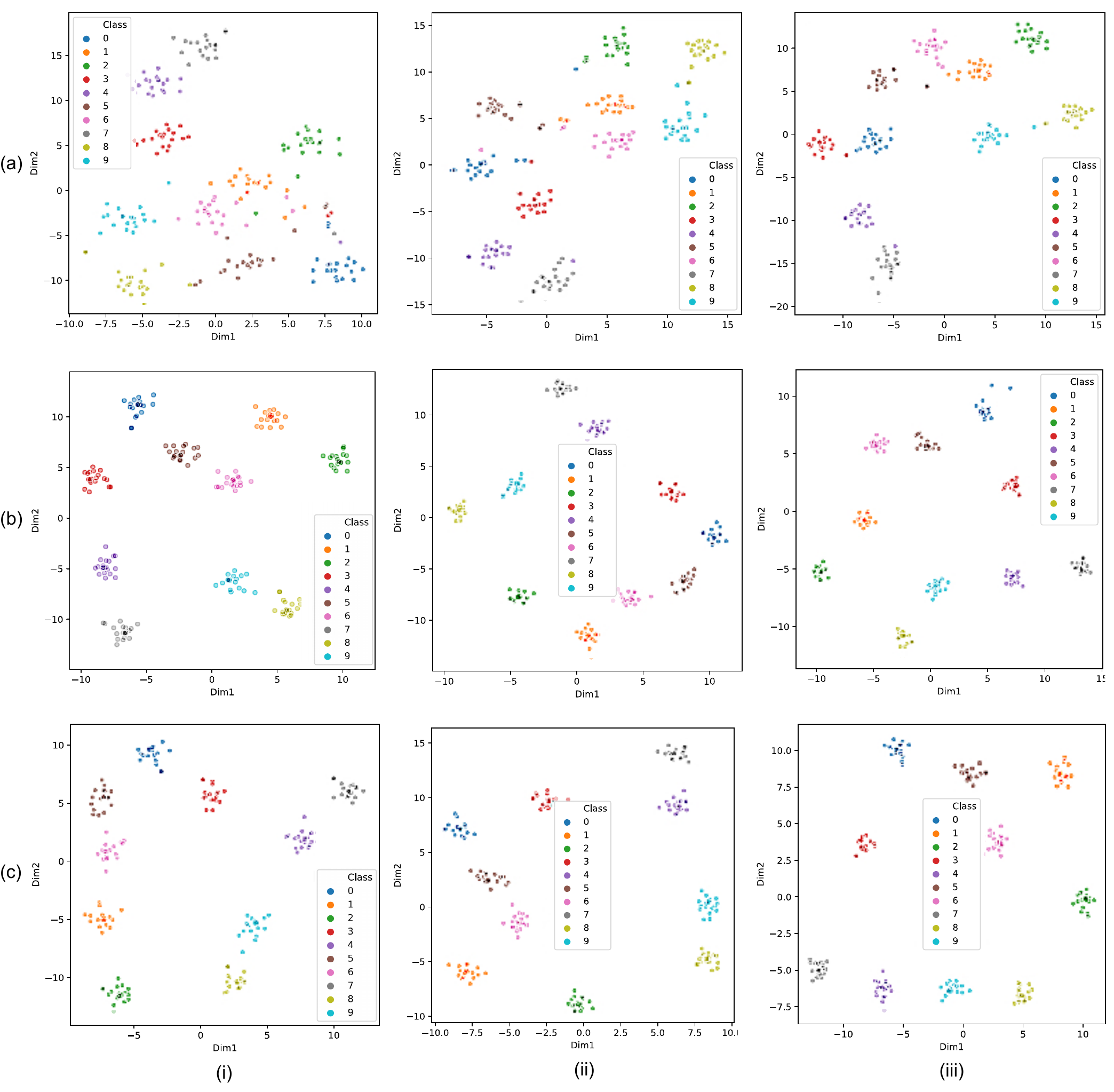}
\caption{tSNE visualization of the feature spaces using the proposed and the best performing benchmarking methods.}
\label{fig:TSNE}
\end{figure*}

To show the discrimination ability of the proposed face recognition methods adopting the GLF-LSTM and SLF-LSTM cell architectures, as well as the best bechchmarking method, i.e. VGG + Conv-LSTM \cite{CSVT}, a t-Distributed Stochastic Neighbor Embedding (tSNE) visualization \cite{tSNE} has been used. The tSNE visualization shows the feature spaces produced by each method in a two dimensional space, thus representing the discrimination ability of the different embeddings. Figure \ref{fig:TSNE} plots the different embeddings feeding the classifier, when \textit{i}) VGG-16 + Conv-LSTM \cite{CSVT} (Figure \ref{fig:TSNE} row (a)), \textit{ii}) ResNet-50+GLF-LSTM (Figure \ref{fig:TSNE} row (b)), and \textit{iii}) ResNet-50+SLF-LSTM (Figure \ref{fig:TSNE} row (c)) are used. To make this visualisation easier to interpret, the t-SNE analysis is performed only for the first 10 subjects available in dataset, respectively for Protocol 1 (Figure \ref{fig:TSNE} column (i)), Protocol 2 (Figure \ref{fig:TSNE} column (ii)), and Protocol 3 (Figure \ref{fig:TSNE} column (iii)) test data. 

The results presented in Figure \ref{fig:TSNE} row (a) clearly show that VGG + Conv-LSTM could not form separate and dense clusters for some test samples. Even for the case of the less challenging Protocol 3, there are still a few samples whose data points are mixed. However, the results presented in the second and third rows of Figure \ref{fig:TSNE}, show that the proposed face recognition methods adopting the novel GLF-LSTM and SLF-LSTM cell architectures create denser and more effective clusters, notably with the data points distributed closer to their centroids. This observation reveals that the subjects are more easily separable in these feature spaces, thus validating the use of the novel cell architectures in the context of the proposed recognition method.

\subsection{Ablation Study}
An ablation study has been performed using the three test protocols, to investigate the influence of each sub-network in the recognition methods in terms of final face recognition performance. Tables \ref{tab: abl1} and \ref{tab: abl2} present the performance of the proposed face recognition methods, respectively adopting the novel GLF-LSTM and SLF-LSTM cell architectures, when removing/changing individual sub-networks from the recognition methods. It should be noted that the CNN feature extraction sub-network should always be kept to extract the essential spatial embeddings. The complete methods' performances are presented in the last row of Tables \ref{tab: abl1} and \ref{tab: abl2}. 

\subsubsection{Impact of joint recurrent learning and attention learning} The performance considering only the CNN feature extraction sub-networks, i.e., without the joint recurrent learning and attention learning sub-networks, is presented in the first row of Tables \ref{tab: abl1} and \ref{tab: abl2}. In this configuration, the spatial embeddings obtained from the horizontal and vertical views are concatenated to be directly used as input to the classifier. This modification significantly degrades the recognition accuracy as it limits the ability to learn the joint relations between the extracted spatial embeddings. Moreover, in the absence of the attention learning sub-network, it is not possible to assign higher weights to the more important recurrently learned embeddings. These results reveal the added value of the joint recurrent learning and attention learning sub-networks in learning a more discriminative spatio-angular model for LF face recognition.

\subsubsection{Impact of bi-directional joint recurrent learning} 
This ablation test changes the joint recurrent learning sub-network by removing the backward connections between GLF-LSTM and SLF-LSTM cells. In this context, the bi-directional LSTM networks become the uni-directional ones, thus only exploiting forward relationships. The results reported in the second row of Tables \ref{tab: abl1} and \ref{tab: abl2} show a mean performance reduction of 0.48\%, showing the added value of adopting a bi-directional joint recurrent learning to exploit both the forward and backward information available in the input sequences. 

\subsubsection{Impact of joint recurrent learning} 
The next ablation experiment removes the whole joint recurrent learning sub-network, thus feeding the attention layer directly with the spatial embeddings obtained by ResNet-50. The results presented in the third row of Tables \ref{tab: abl1} and \ref{tab: abl2} show a performance degradation of 3.31\% and 3.21\% when removing GLF-LSTM and SLF-LSTM bi-directional networks, respectively. This demonstrates the benefit of using the joint recurrent learning sub-network in learning the relationships between spatial embeddings.  

\subsubsection{Impact of attention learning} 
When the attention learning sub-network is removed, the recurrently learned spatio-angular embeddings maintain their weights for classification, instead of selectively focusing on the most important LSTM embeddings. The results in the fourth and fifth rows of Tables \ref{tab: abl1} and \ref{tab: abl2} show performance degradations when GLF and SLF bi-directional and uni-directional LSTM networks have been used without an attention learning layer.

\begin{table}[!t]
\centering
\caption{Ablation study for the ResNet-50 + GLF-LSTM face recognition method.}
\setlength
\tabcolsep{4pt}
\begin{tabular}{ l| l| l| l| l| l| l| l}
\hline
    \textbf{CNN} & \textbf{LSTM} & \textbf{Bi-LSTM} & \textbf{Att.} & \textbf{Prot. 1} & \textbf{Prot. 2} & \textbf{Prot. 3} & \textbf{Mean}\\ \hline\hline
    \cmark & \xmark & \xmark & \xmark &  90.15 & 94.15  & 97.50  & 93.93 \\
    \cmark & \cmark & \xmark & \cmark &  96.90 & 97.80  & 99.40  & 98.03 \\
    \cmark & \xmark & \xmark & \cmark & 92.40  &  95.40 &  97.80  & 95.20 \\
    \cmark & \cmark & \xmark & \xmark & 94.80 &  96.65 & 99.50 & 96.98 \\
    \cmark & \xmark & \cmark & \xmark & 95.70  &  97.15 & 99.40  & 97.41 \\
    \cmark & \xmark & \cmark & \cmark & \textbf{97.75}  & \textbf{98.00}  & \textbf{99.80}  & \textbf{98.51} \\
    \hline
\end{tabular}
\label{tab: abl1}
\end{table}

\begin{table}[!t]
\centering
\caption{Ablation study for the ResNet-50 + SLF-LSTM face recognition method.}
\setlength
\tabcolsep{4pt}
\begin{tabular}{ l| l| l| l| l| l| l| l}
\hline
    \textbf{CNN} & \textbf{LSTM} & \textbf{Bi-LSTM} & \textbf{Att.} & \textbf{Prot. 1} & \textbf{Prot. 2} & \textbf{Prot. 3} & \textbf{Mean}\\ \hline\hline
    \cmark & \xmark & \xmark & \xmark &  90.15 &  94.15 &  97.50 & 93.93 \\
    \cmark & \cmark & \xmark & \cmark & 96.90  &  97.60 & 99.30  & 97.93 \\
    \cmark & \xmark & \xmark & \cmark &  92.40 & 95.40  &  98.30 & 95.20 \\
    \cmark & \cmark & \xmark & \xmark & 94.50 & 96.70  & 99.40 & 96.86 \\
    \cmark & \xmark & \cmark & \xmark & 95.65 & 97.00 &  99.40 & 97.36 \\
    \cmark & \xmark & \cmark & \cmark & \textbf{97.65} & \textbf{98.00}  & \textbf{99.60}  & \textbf{98.41} \\
    \hline
\end{tabular}
\label{tab: abl2}
\end{table}

\section{Summary and Future Work}
This paper proposes two novel LSTM cell architectures able to jointly learn a model from multiple dependent sequences of input data. The novel LSTM cell architectures adopt gate-level fusion and state-level fusion to create richer joint embeddings to be used for visual analysis tasks. To show the efficiency of the novel LSTM cell architectures, they have been integrated into deep learning-based methods for face recognition with LF images. The resulting face recognition methods jointly learn the scene horizontal and vertical parallaxes, capturing richer spatio-angular information from these directions. A comprehensive evaluation has been conducted on the IST-EURECOM LFFD dataset using three challenging test protocols. The obtained performance results show the superiority of the proposed face recognition methods based on the novel LSTM cell architectures over six state-of-the-art benchmarking methods. Additionally, the novel LSTM cell architectures have been compared when using different fusion strategies to show the benefits of the novel architectures in terms of learning the dependencies between the  input sequences for LF-based face recognition.

The proposed LSTM cell architectures have been used for LF-based face recognition in the context of this paper. The proposed LSTM cell architectures can be generic enough to address different visual recognition tasks, as accepting as input two or more related sequences. Hence, future work will consider the use of these LSTM cell architectures for other recognition tasks, notably for multiple dependent and synchronized sequences such as video streams from multiple synchronized cameras or audio and visual sequences from movie clips.

\bibliographystyle{ieeetran}
\bibliography{main}

\begin{thebibliography}{10}
\providecommand{\url}[1]{#1}
\csname url@samestyle\endcsname
\providecommand{\newblock}{\relax}
\providecommand{\bibinfo}[2]{#2}
\providecommand{\BIBentrySTDinterwordspacing}{\spaceskip=0pt\relax}
\providecommand{\BIBentryALTinterwordstretchfactor}{4}
\providecommand{\BIBentryALTinterwordspacing}{\spaceskip=\fontdimen2\font plus
\BIBentryALTinterwordstretchfactor\fontdimen3\font minus
  \fontdimen4\font\relax}
\providecommand{\BIBforeignlanguage}[2]{{%
\expandafter\ifx\csname l@#1\endcsname\relax
\typeout{** WARNING: IEEEtran.bst: No hyphenation pattern has been}%
\typeout{** loaded for the language `#1'. Using the pattern for}%
\typeout{** the default language instead.}%
\else
\language=\csname l@#1\endcsname
\fi
#2}}
\providecommand{\BIBdecl}{\relax}
\BIBdecl

\bibitem{LeCun2}
Y.~LeCun, Y.~Bengio, and G.~Hinton, ``Deep learning,'' \emph{Nature}, vol. 521,
  no. 7553, pp. 436--444, May 2015.

\bibitem{deepsurvey}
J.~Schmidhuber, ``Deep learning in neural networks: An overview,'' \emph{Neural
  networks}, vol.~61, no.~1, pp. 85--117, January 2015.

\bibitem{LeCun}
Y.~{LeCun}, ``Deep learning hardware: Past, present, and future,'' in
  \emph{International Solid- State Circuits Conference}, San Francisco, CA,
  USA, February 2019.

\bibitem{CNNSurvey}
W.~Liu, Z.~Wang, X.~Liu, N.~Zeng, Y.~Liu, and F.~E. Alsaadi, ``A survey of deep
  neural network architectures and their applications,'' \emph{Neurocomputing},
  vol. 234, no.~1, pp. 11 -- 26, April 2017.

\bibitem{RNNSurvey}
Z.~C. Lipton, J.~Berkowitz, and C.~Elkan, ``A critical review of recurrent
  neural networks for sequence learning,'' \emph{arXiv preprint
  arXiv:1506.00019}, 2015.

\bibitem{LSTM}
S.~Hochreiter and J.~Schmidhuber, ``Long short-term memory,'' \emph{Neural
  computation}, vol.~9, no.~8, pp. 1735--1780, November 1997.

\bibitem{LSTMOD}
K.~Greff, R.~K. Srivastava, J.~Koutn{\'\i}k, B.~R. Steunebrink, and
  J.~Schmidhuber, ``{LSTM}: A search space odyssey,'' \emph{IEEE transactions
  on neural networks and learning systems}, vol.~28, no.~10, pp. 2222--2232,
  July 2016.

\bibitem{SPL}
X.~{Wang}, L.~{Gao}, J.~{Song}, and H.~{Shen}, ``Beyond frame-level {CNN}:
  Saliency-aware {3-D} {CNN} with {LSTM} for video action recognition,''
  \emph{IEEE Signal Processing Letters}, vol.~24, no.~4, pp. 510--514, April
  2017.

\bibitem{act3}
A.~{Ullah}, J.~{Ahmad}, K.~{Muhammad}, M.~{Sajjad}, and S.~W. {Baik}, ``Action
  recognition in video sequences using deep {Bi-Directional LSTM With CNN}
  features,'' \emph{IEEE Access}, vol.~6, no.~1, pp. 1155--1166, November 2017.

\bibitem{CSVT}
A.~Sepas-Moghaddam, P.~Correia, K.~Nasrolahi, T.~Moeslund, and F.~Pereira, ``A
  double-deep spatio-angular learning framework for light field based face
  recognition,'' \emph{IEEE Transactions on Circuits and Systems for Video
  Technology}, vol. in press, March 2020.

\bibitem{faceLSTM}
S.~Gong, Y.~Shi, A.~K. Jain, and N.~D. Kalka, ``Recurrent embedding aggregation
  network for video face recognition,'' \emph{arXiv preprint arXiv:1904.12019},
  2019.

\bibitem{ACII}
A.~{Sepas-Moghaddam}, A.~{Etemad}, P.~{Correia}, and F.~{Pereira}, ``A deep
  framework for facial emotion recognition using light field images,'' in
  \emph{International Conference on Affective Computing and Intelligent
  Interaction}, Cambridge, UK, September 2019.

\bibitem{emotLSTM}
Y.~Fan, X.~Lu, D.~Li, and Y.~Liu, ``Video-based emotion recognition using
  {CNN-RNN and C3D} hybrid networks,'' in \emph{International Conference on
  Multimodal Interaction}, Tokyo, Japan, October 2016, pp. 445--450.

\bibitem{lip}
A.~Garg, J.~Noyola, and S.~Bagadia, ``Lip reading using {CNN and LSTM},''
  Technical report, Stanford University, CS231n project report, Tech. Rep.,
  2016.

\bibitem{LSTMdesc}
J.~{Donahue}, L.~A. {Hendricks}, M.~{Rohrbach}, S.~{Venugopalan},
  S.~{Guadarrama}, K.~{Saenko}, and T.~{Darrell}, ``Long-term recurrent
  convolutional networks for visual recognition and description,'' \emph{IEEE
  Transactions on Pattern Analysis and Machine Intelligence}, vol.~39, no.~4,
  pp. 677--691, September 2017.

\bibitem{fuse}
A.~Ross and A.~Jain, ``Information fusion in biometrics,'' \emph{Pattern
  recognition letters}, vol.~24, no.~13, pp. 2115--2125, September 2003.

\bibitem{audiovisual}
E.~Avots, T.~Sapi{\'n}ski, M.~Bachmann, and D.~Kami{\'n}ska, ``Audiovisual
  emotion recognition in wild,'' \emph{Machine Vision and Applications},
  vol.~30, no.~5, pp. 975--985, July 2019.

\bibitem{lensletLF}
R.~Ng, M.~Levoy, M.~Bradif, G.~Duval, M.~Horowitz, and P.~Hanrahan, ``Light
  field photography with a hand-held plenoptic camera,'' \emph{Tech Report CSTR
  2005-02}, February 2005.

\bibitem{LF}
M.~Levoy and P.~Hanrahan, ``Light field rendering,'' in \emph{Annual Conference
  on Computer Graphics and Interactive Techniques}, New York, NY, USA, August
  1996.

\bibitem{ryrb13}
R.~Raghavendra, B.~Yang, K.~Raja, and C.~Busch, ``A new perspective: Face
  recognition with light-field camera,'' in \emph{International Conference on
  Biometrics}, Madrid, Spain, June 2013.

\bibitem{rryb13}
R.~Raghavendra, K.~Raja, B.~Yang, and C.~Busch, ``Multi-face recognition at a
  distance using light-field camera,'' in \emph{International Conference on
  Intelligent Information Hiding and Multimedia Signal Processing}, Beijing,
  China, July 2013.

\bibitem{rrb16}
R.~Raghavendra, K.~Raja, and C.~Busch, ``Exploring the usefulness of light
  field cameras for biometrics: An empirical study on face and iris
  recognition,'' \emph{IEEE Transactions on Information Forensics and
  Security}, vol.~11, no.~5, pp. 922--936, May 2016.

\bibitem{icip}
A.~Sepas-Moghaddam, P.~Correia, and F.~Pereira, ``Light field local binary
  patterns description for face recognition,'' in \emph{International
  Conference on Image Processing}, Beijing, China, September 2017.

\bibitem{ear}
A.~Sepas-Moghaddam, F.~Pereira, and P.~Correia, ``Ear recognition in a light
  field imaging framework: A new perspective,'' \emph{IET Biometrics}, vol.~7,
  no.~3, pp. 224--231, May 2018.

\bibitem{MLSP}
A.~Sepas-Moghaddam, P.~Correia, K.~Nasrollahi, T.~Moeslund, and F.~Pereira,
  ``Light field based face recognition via a fused deep representation,'' in
  \emph{International Workshop on Machine Learning for Signal Processing},
  Aalborg, Denmark, September 2018.

\bibitem{TIFS}
A.~{Sepas-Moghaddam}, F.~{Pereira}, and P.~L. {Correia}, ``Light field based
  face presentation attack detection: Reviewing, benchmarking and one step
  further,'' \emph{IEEE Transactions on Information Forensics and Security},
  vol.~13, no.~7, pp. 1696--1709, July 2018.

\bibitem{IETanti}
A.~Sepas-Moghaddam, L.~Malhadas, P.~Correia, and F.~Pereira, ``Face spoofing
  detection using a light field imaging framework,'' \emph{IET Biometrics},
  vol.~7, no.~1, pp. 39--48, January 2018.

\bibitem{rrb15}
R.~Raghavendra, K.~Raja, and C.~Busch, ``Presentation attack detection for face
  recognition using light field camera,'' \emph{IEEE Transactions on Image
  Processing}, vol.~24, no.~3, pp. 1060--1075, March 2015.

\bibitem{ICASSP}
A.~Sepas-Moghaddam, A.~Etemad, P.~Correia, and F.~Pereira, ``Facial emotion
  recognition using light field images with deep attention-based bidirectional
  {LSTM},'' in \emph{International Conference on Acoustics, Speech, and Signal
  Processing}, Barcelona, Spain, May 2020.

\bibitem{pvz15}
O.~Parkhi, A.~Vedaldi, and A.~Zisserman, ``Deep face recognition,'' in
  \emph{British Machine Vision Conference}, Swansea, UK, September 2015.

\bibitem{iwbf}
A.~Sepas-Moghaddam, V.~Chiesa, P.~Correia, F.~Pereira, and J.~Dugelay, ``The
  {IST-EURECOM} light field face database,'' in \emph{International Workshop on
  Biometrics and Forensics}, Coventry, UK, April 2017.

\bibitem{backpro}
P.~J. {Werbos}, ``Backpropagation through time: what it does and how to do
  it,'' \emph{Proceedings of the IEEE}, vol.~78, no.~10, pp. 1550--1560,
  October 1990.

\bibitem{vanish}
S.~Hochreiter, ``The vanishing gradient problem during learning recurrent
  neural nets and problem solutions,'' \emph{International Journal of
  Uncertainty, Fuzziness and Knowledge-Based Systems}, vol.~6, no.~2, pp.
  107--116, April 1998.

\bibitem{peep}
F.~A. {Gers} and J.~{Schmidhuber}, ``Recurrent nets that time and count,'' in
  \emph{IEEE-INNS-ENNS International Joint Conference on Neural Networks},
  Como, Italy, July 2000.

\bibitem{bilstm}
A.~Graves and J.~Schmidhuber, ``Framewise phoneme classification with
  bidirectional lstm and other neural network architectures,'' \emph{Neural
  networks}, vol.~18, no.~5, pp. 602--610, August 2005.

\bibitem{ab91}
E.~Adelson and J.~Bergen, ``The plenoptic function and the elements of early
  vision,'' in \emph{Computation Models of Visual Processing}.\hskip 1em plus
  0.5em minus 0.4em\relax Cambridge MA, USA: MIT Press, 1991, pp. 3--20.

\bibitem{lumi}
S.~Gortler, R.~Grzeszczuk, R.~Szeliski, and M.~Cohen, ``The lumigraph,'' in
  \emph{Annual Conference on Computer Graphics and Interactive Techniques}, New
  Orleans, LA, USA, August 1996.

\bibitem{d14}
D.~Dansereau, ``Plenoptic signal processing for robust vision in field
  robotics,'' Ph.D. dissertation, Mechatronic Engineering, Queensland
  University of Technology, Queensland, Australia, January 2014.

\bibitem{LFsurvey}
C.~Galdi, V.~Chiesa, C.~Busch, P.~Correia, J.-L. Dugelay, and C.~Guillemot,
  ``Light fields for face analysis,'' \emph{Sensors}, vol.~19, no.~12, p. 2687,
  June 2019.

\bibitem{lpv08}
H.~Lu, K.~Plataniotis, and A.~Venetsanopoulos, ``{MPCA}: Multilinear principal
  component analysis of tensor objects,'' \emph{IEEE Transactions on Neural
  Networks}, vol.~19, no.~1, pp. 18--39, January 2008.

\bibitem{rryb16}
R.~{Raghavendra}, K.~B. {Raja}, B.~{Yang}, and C.~{Busch}, ``Comparative
  evaluation of super-resolution techniques for multi-face recognition using
  light-field camera,'' in \emph{International Conference on Digital Signal
  Processing}, Fira, Greece, July 2013.

\bibitem{resnet}
K.~He, X.~Zhang, S.~Ren, and J.~Sun, ``Deep residual learning for image
  recognition,'' in \emph{International conference on computer vision and
  pattern recognition}, Las Vegas, NV, USA, June 2016.

\bibitem{LFTool}
D.~Dansereau. Light field toolbox v. 0.4, [online]. available:
  https://github.com/doda42/lftoolbox/, [accessed {May} 2020].

\bibitem{csxpz18}
Q.~Cao, L.~Shen, W.~Xie, M.~Parkhi, and A.~Zisserman, ``{VGGFace2}: A dataset
  for recognising faces across pose and age,'' in \emph{International
  Conference on Automatic Face and Gesture Recognition}, Xi'an, China, May 2018.

\bibitem{seresnet}
J.~Hu, L.~Shen, and G.~Sun, ``Squeeze-and-excitation networks,'' in
  \emph{computer vision and pattern recognition}, Salt Lake City, UT, USA, June
  2018.

\bibitem{att4}
T.~Rocktäschel and E.~Grefenstette, ``Reasoning about entailment with neural
  attention,'' in \emph{International Conference on Learning Representations},
  San Juan, Puerto Rico, May 2016.

\bibitem{tensorflow}
M.~Abadi, P.~Barham, and J.~Chen, ``Tensorflow: A system for large-scale
  machine learning,'' in \emph{International Symposium on Operating Systems
  Design and Implementation}, Savannah, GA, USA, November 2016.

\bibitem{keras}
F.~Chollet, ``Keras,'' \url{https://keras.io}, 2015.

\bibitem{Lytro}
Lytro website, \textit{Lytro Inc, 2016.} [online]. available:
  https://www.lytro.com, [accessed {December} 2019].

\bibitem{oneshot}
{Li Fei-Fei}, R.~{Fergus}, and P.~{Perona}, ``One-shot learning of object
  categories,'' \emph{IEEE Transactions on Pattern Analysis and Machine
  Intelligence}, vol.~28, no.~4, pp. 594--611, February 2006.

\bibitem{oneshotface}
L.~{Wang}, Y.~{Li}, and S.~{Wang}, ``Feature learning for one-shot face
  recognition,'' in \emph{IEEE International Conference on Image Processing},
  Athens, Greece, September 2018.

\bibitem{DLBP}
A.~{Aissaoui}, J.~{Martinet}, and C.~{Djeraba}, ``{DLBP}: A novel descriptor
  for depth image based face recognition,'' in \emph{IEEE International
  Conference on Image Processing}, Paris, France, October 2014.

\bibitem{LSTMact2}
J.~{Liu}, A.~{Shahroudy}, D.~{Xu}, A.~C. {Kot}, and G.~{Wang}, ``Skeleton-based
  action recognition using spatio-temporal {LSTM} network with trust gates,''
  \emph{IEEE Transactions on Pattern Analysis and Machine Intelligence},
  vol.~40, no.~12, pp. 3007--3021, November 2018.

\bibitem{tSNE}
L.~v.~d. Maaten and G.~Hinton, ``Visualizing data using {t-SNE},''
  \emph{Journal of machine learning research}, vol.~9, no.~1, pp. 2579--2605,
  November 2008.

\end{thebibliography}

\end{document}